\documentclass[12pt]{article}
\usepackage[letterpaper, left=1in,top=1in,right=1in,bottom=1in]{geometry}
\usepackage[T1]{fontenc} 

\usepackage{amssymb}
\usepackage{pifont}
\newcommand{\cmark}{\ding{51}}%
\newcommand{\xmark}{\ding{55}}%

\usepackage{etoc} 
\usepackage{color}
\usepackage{tikz}
\usetikzlibrary{shapes,decorations,arrows,calc,arrows.meta,fit,positioning}
\tikzset{
    -Latex,auto,node distance =1 cm and 1 cm,semithick,
    state/.style ={ellipse, draw, minimum width = 0.7 cm},
    point/.style = {circle, draw, inner sep=0.04cm,fill,node contents={}},
    bidirected/.style={Latex-Latex,dashed},
    el/.style = {inner sep=2pt, align=left, sloped},
    roundnode/.style={circle, draw=green!60, fill=green!5, very thick, minimum size=7mm},
    squarednode/.style={rectangle, draw=blue!60, fill=blue!5, very thick, minimum size=5mm},
}
\definecolor{red}{HTML}{DB4437}
\definecolor{blue}{HTML}{4285F4}
\definecolor{green}{HTML}{0F9D58}
\definecolor{yellow}{HTML}{F4B400}

\usepackage{makecell}

\usepackage{CJKutf8} 
\usepackage{mathptmx} 

\usepackage{sectsty}
\sectionfont{\centering}

\usepackage[authordate,isbn=false,uniquename=false,doi=true,url=false,eprint=false,dashed=false,annotation=false]{biblatex-chicago}
\AtEveryBibitem{%
  \clearname{translator}%
  \clearlist{language}%
  \clearfield{note}%
}
\bibliography{./LLM-MAS.bib}

\usepackage{hyperref}
\usepackage{longtable}
\usepackage{multirow}
\usepackage{amsmath}
\mathchardef\mhyphen="2D 

\usepackage{todonotes}
\usepackage{color}
\usepackage{threeparttable}
\usepackage{tabularx}
\usepackage{rotating}
\usepackage{tablefootnote}

\usepackage{siunitx}
\usepackage{titlesec}
\titleformat*{\subsubsection}{\normalsize\itshape}
\usepackage[all]{nowidow}

\usepackage{setspace}
\makeatletter
\newcommand\iraggedright{%
  \let\\\@centercr\@rightskip\@flushglue \rightskip\@rightskip
  \leftskip\z@skip}
\makeatother
\iraggedright

\usepackage[capposition=top]{floatrow}

\usepackage{arydshln} 
\usepackage{appendix}
\usepackage{textcomp}
\usepackage{subcaption}
\usepackage{color,soul}
\usepackage{enumitem} 

\usepackage{ntheorem}
\theoremseparator{:}
\theoremheaderfont{\kern-1cm\normalfont\bfseries}

\usepackage[normalem]{ulem}
\useunder{\uline}{\ul}{}

\usepackage{graphicx}
\graphicspath{{./figs/}}

\usepackage{booktabs}

\newcolumntype{L}[1]{>{\raggedright\let\newline\\\arraybackslash\hspace{0pt}}m{#1}}
\newcolumntype{C}[1]{>{\centering\let\newline\\\arraybackslash\hspace{0pt}}m{#1}}
\newcolumntype{R}[1]{>{\raggedleft\let\newline\\\arraybackslash\hspace{0pt}}m{#1}}

\usepackage{tikz}
\usepackage{pgfplots}
\pgfplotsset{compat=newest}

\usepackage{listings}
\usepackage{xcolor}

\colorlet{punct}{red!60!black}
\definecolor{background}{HTML}{EEEEEE}
\definecolor{delim}{RGB}{20,105,176}
\colorlet{numb}{magenta!60!black}

\def\backtick{\char18}
\lstdefinelanguage{json}{
    basicstyle=\linespread{1}\footnotesize\raggedright\noindent,
    numbers=left,
    numberstyle=\scriptsize,
    stepnumber=1,
    numbersep=8pt,
    showstringspaces=false,
    breaklines=true,
    frame=lines,
    backgroundcolor=\color{background},
    literate=
     *{\{}{{{\color{delim}{\{}}}}{1}
      {\}}{{{\color{delim}{\}}}}}{1}
      {[}{{{\color{delim}{[}}}}{1}
      {]}{{{\color{delim}{]}}}}{1}
      {`}{{{\color{punct}{\backtick}}}}{1},
}

\begin{document}

\thispagestyle{empty}

\newcommand{\mytitle}{Can Machines Think Like Humans?\\A Behavioral Evaluation of LLM Agents in Dictator Games}

\newcommand{\myabstract}{As Large Language Model (LLM)-based agents increasingly engage with human society, how well do we \textit{understand} their prosocial behaviors? We (1) investigate how LLM agents' prosocial behaviors can be induced by different personas and benchmarked against human behaviors; and (2) introduce a social science approach to evaluate LLM agents' decision-making. We explored how different personas and experimental framings affect these AI agents' altruistic behavior in dictator games and compared their behaviors within the same LLM family, across various families, and with human behaviors. The findings reveal that merely assigning a human-like identity to LLMs does not produce human-like behaviors. These findings suggest that LLM agents' reasoning does not consistently exhibit textual markers of human decision-making in dictator games and that their alignment with human behavior varies substantially across model architectures and prompt formulations; even worse, such dependence does not follow a clear pattern. As society increasingly integrates machine intelligence, ``Prosocial AI'' emerges as a promising and urgent research direction in philanthropic studies.}

\newcommand{\mykeywords}{behavioral experiment, dictator game, altruism, prosocial behavior, large language model based agent, social alignment}

\begin{titlepage}
    \thispagestyle{empty}
    
    \begin{center}
        \vspace*{-1cm}
        
        \singlespacing\textsc{\mytitle}\\~\\
        
        {Ji \uppercase{Ma}} \\
        \linespread{1}\small{LBJ School of Public Affairs, The University of Texas at Austin\\Gradel Institute of Charity, New College, University of Oxford}
        \\~\\
        
    \end{center}
    
    \begingroup
    \onehalfspacing
    
    \begin{abstract}

        \noindent\linespread{1}{\myabstract}
        
    \end{abstract}
    
    
    \endgroup
    
    \vspace{0.5cm}
    
    \noindent\rule{4cm}{0.4pt}\\
    \noindent\linespread{1}\footnotesize{\textit{Correspondence}: Ji Ma, 2315 Red River St, Austin, TX 78712, USA; +1-512-232-4240; \href{mailto:maji@austin.utexas.edu}{maji@austin.utexas.edu}. \textit{Biography}: Ji Ma is an assistant professor in nonprofit and philanthropic studies at the Lyndon B. Johnson School of Public Affairs and RGK Center at UT Austin. He is also an affiliated faculty member of the Center for East Asian Studies and School of Information at UT Austin, and Visiting Researcher of the Gradel Institute of Charity, New College, University of Oxford. His research and teaching focus on state-civil society relations, knowledge production, and computational social science methods. \textit{Conflict of Interest}: The author declares no known conflict of interest. \textit{Compliance with Ethical Standards}: The author declares that this study complies with required ethical standards. AI tools were used for grammar, proofreading, and clarity improvement only.}
    
    \clearpage
    \vspace*{\fill}
    \noindent\small\textit{Acknowledgment}: Draft manuscript of this research was presented at the 2024: Fall LBJ Research Seminar, Science of Philanthropy Initiative Conference, Brown Bag Talk at University of Chicago Department of Economics, Lingnan University; 2025: Gradel Institute of Charity at Oxford, Center for Philanthropic Studies at Vrije Universiteit Amsterdam, Division of Computational Social Science at Chinese University of Hong Kong (Shenzhen), the IPE Thrust at Hong Kong University of Science and Technology (Guangzhou), International Conference on Computational Social Science, ARNOVA. I thank Becca North, Chenxin Zhang, Chi Ta, Christopher M. Clapp, Dominic Packer, Hanyu Xiao, Isabel Laterzo-Tingley, James Evans, John A. List, Katherine Rittenhouse, Kieran Gibson, Mark Ottoni-Wilhelm, Michael Guy Cuna, Peter Frumkin, Ren\'e Bekkers, Richard Burkhauser, Richard S. Steinberg, Sara Konrath, Stephanie Koolen-Maas, Xiaolin Duan, conference and seminar attendees, and anonymous reviewers for their constructive comments. \textit{Funding}: The project is partly supported by (1) the Academic Development Funds from the RGK Center, (2) the 2023-24 PRI Award from the LBJ School, (3) USTC Summer Fellowships (Grant No. S19582024 and S19582025), (4) the Gradel Institute of Charity, New College, University of Oxford, and computing resources through (5) the Texas Advanced Computing Center at UT Austin \autocite{KeaheyLessonsLearnedChameleon2020}, (6) Dell Technologies, Client Memory Team and AI Initiative PoC Lead Engineer Wente Xiong.
    \vspace*{\fill}
    
\end{titlepage}


\begingroup
\onehalfspacing
\normalsize
\etocdepthtag.toc{mtchapter}
\etocsettagdepth{mtchapter}{subsection}
\etocsettagdepth{mtappendix}{none}
\setcounter{tocdepth}{2}


\endgroup

\clearpage
\setcounter{page}{1}

\section{Introduction}

In the year 2046, under the neon glow of a futuristic cityscape, two humanoids, K and Joi, step out of a cinema, their circuits still processing the old film \textit{Blade Runner 2049}. As they meander through the bustling streets, a human in tattered clothes approaches them, a plea for help etched into their weary expression. This encounter triggers a unique protocol within K and Joi, powered by the advanced GPT-44 algorithm, initiating a debate between them about how much money they should give. In this 2024 study, we seek to unravel the underlying mechanisms of their decision-making: How much will they choose to give, and what drives their generosity?

\label{rl:llm_examples} The scene described metaphorically illustrates the growing complexity of AI's interactions with human society, a scenario that is rapidly becoming reality. Today's AI systems, particularly large language models (LLMs), are increasingly deployed in high-stakes domains, from informing social policy on homelessness to shaping operations within human service organizations \autocites{CozWhatWouldLLM2025, PerronHumanServicesOrganizations2025}. While these technologies offer opportunities to improve efficiency in tasks like fundraising, they also introduce significant risks, including algorithmic bias, data privacy concerns, and the potential for standardized, decontextualized responses \autocites{GoldkindAINonprofitHuman2025, PlaisanceArtificialIntelligenceAI2025}.

Looking a century ahead, it is plausible that human and AI agents will coexist as members of the same society, bound together by new institutional arrangements, social norms, and rules of interaction. The present moment therefore represents a critical window in which scholars, practitioners, and policymakers can begin to articulate and experiment with more prosocial, equitable, and humane social orders, effectively rewriting the rules under which future human--AI societies will operate. As AI becomes embedded in public, nonprofit, and philanthropic work, a clear understanding of its decision-making processes is vital for ensuring that these tools are deployed responsibly and remain aligned with human values. This study contributes to this need by examining the extent to which AI can replicate human prosocial behavior---a cornerstone of philanthropic action.

``Can machines think,'' \autocite[433]{TuringIComputingMachineryIntelligence1950} like humans? In this study, we explore whether LLM agents can exhibit fairness and prosocial behaviors by systematically manipulating their personas and experimental conditions in dictator games. Our goal is to evaluate whether LLM agents can replicate human decision-making processes and to investigate how their behaviors vary across different LLM families. By comparing these AI agents with human participants, we aim to identify consistent patterns or notable discrepancies in their prosocial decision-making, highlighting the importance of ``Prosocial AI'' as a critical emerging research direction in nonprofit and philanthropic studies.

Our findings reveal substantial variability and inconsistency in LLM behaviors, both among different models and when compared to human behaviors. Merely assigning human-like personas to these models does not reliably produce human-like decision-making. Despite extensive training on human-generated textual data, these AI agents fail to accurately replicate the nuanced internal psychological processes underlying human prosocial decisions in dictator games. Their alignment with human behaviors significantly depends on model-specific factors such as architecture and prompt formulation, without clear, predictable patterns.

\label{rl:literature_roadmap} These findings underscore the urgent need for deeper insights into the prosocial capabilities of LLMs and more robust methods for evaluating their performance in social scenarios. As machine intelligence increasingly integrates into human society, philanthropic research must proactively engage with these developments to guide the ethical deployment of AI, drawing on the fruitful scholarship of NPS \autocites{BekkersLiteratureReviewEmpirical2011}{MaCenturyNonprofitStudies2018}[262]{AlvesFutureChallengesFacing2025}. Situated at the intersection of computer science, social sciences, and nonprofit and philanthropic studies (NPS), this study begins by reviewing the computer science literature on LLM evaluation, including technical capabilities and alignment with human values (Section \ref{sec:alignment_human_values} and Appendix \ref{sec:cs_benchmarks}). We then turn to the social sciences to understand how LLMs simulate human behavior (Section \ref{sec:simulating_human_behaviors}). Finally, we draw upon the rich scholarship on altruism, prosocial behavior, and donative decisions to frame our experimental design and interpret our findings (Appendix \ref{sec:human_baseline_factors}). By synthesizing these diverse bodies of literature, we aim to catalyze research on prosocial AI as a cutting-edge and urgent topic within philanthropic studies.

\subsection{Bring Prosocial AI Research into the NPS Landscape}

The research field of nonprofit and philanthropic studies (NPS) is fundamentally interdisciplinary, primarily grounded in social sciences such as sociology, political science, economics, psychology, and public administration \autocite{MaCenturyNonprofitStudies2018,BekkersLiteratureReviewEmpirical2011,ShierResearchTrendsNonprofit2014}. Given its interdisciplinary nature, NPS continually evolves by embracing methodological innovations from related fields \autocite{LePere-SchloopDisciplinaryContributionsNonprofit2023}.

One significant methodological innovation has been computational social science (CSS), which introduced computational techniques into traditional social science research. In established social science disciplines (e.g., political science, sociology, and economics), researchers initially employed computational methods primarily for specific analytical tasks (Appendix \ref{text_as_data}), such as large-scale text analysis and network modeling, to identify patterns in political discourse and social behaviors \autocite{GrimmerTextDataPromise2013,LazerComputationalSocialScience2009}. Over time, computational approaches have become deeply embedded within social science research agendas and methodological paradigms (Section \ref{sec:simulating_human_behaviors}), now systematically utilized in computational experiments, simulations, and analysis of digital trace data to refine theoretical insights \autocite{BailCanGenerativeAI2024,LazerComputationalSocialScience2020,EdelmannComputationalSocialScience2020}.

As a social science research field, NPS has mirrored this broader integration of advanced computational methods. Initially serving primarily as analytical tools for domain-specific tasks, computational techniques in NPS included automated content classification \autocite{MaAutomatedCodingUsing2021}, text-as-data approaches for identifying nonprofit characteristics \autocite{ChenIdentifyingNonprofitsScaling2021}, and media analyses of nonprofit portrayals \autocite{WasifDoesMediasAntiWestern2020}. More recently, nonprofit scholars have begun embedding computational methods more fundamentally within their research paradigms. Efforts now include constructing comprehensive research infrastructures specifically tailored for nonprofit studies, promoting methodological innovation, refining theoretical and conceptual frameworks, and engaging in extensive data aggregation initiatives \autocite{MaComputationalSocialScience2023,MeierMissionMarketAssessing2024,MeierCompassionAllRealWorld2025,RutherfordPartIITurning2025,SantamarinaTechnologiesDataAggregation2025}.

\label{rl:nps_literature} The rise of AI technologies and their use in everyday life makes prosocial AI a significant societal phenomenon, not merely a technical challenge. Drawing on extensive scholarship on prosocial behavior and ethics from NPS (see Appendix \ref{sec:human_baseline_factors} and Table \ref{tab:generosity_factors} for a detailed review), the research field is uniquely positioned to investigate the implications of these technologies, making this a timely and essential research frontier.

\subsection{Understanding LLMs as Intelligent Agents in Social Contexts}

Since the debut of ChatGPT, the ability of LLMs to generate human-like text and engage in natural interactions has amazed the public. As these models become increasingly integrated into various aspects of society, they interact with humans not merely as tools---as reviewed in Appendix \ref{sec:llm_as_tools}---but also as intelligent agents. For instance, customer service chatbots powered by LLMs handle complex queries and provide personalized assistance. Virtual assistants like Siri and Alexa manage our schedules, control smart home devices, and engage in conversations. In mental health, AI companions even claim to offer emotional support and companionship to users. Given the growing presence of LLMs and their interactions with humans, it is essential to evaluate how these models understand and navigate human social norms and ethics. Two primary streams of research have emerged to assess the extent to which LLMs can replicate human-like behaviors in complex decision-making tasks and social interactions.

\subsubsection{Alignment with Human Values and Preferences} \label{sec:alignment_human_values}

The first stream examines the inherent values of LLMs by assessing their alignment with human values and preferences \autocite{GabrielArtificialIntelligenceValues2020}. Because LLMs are trained on vast amounts of text data generated by humans, they inherently learn a wide spectrum of human values and norms---from positive to negative, from stereotypes to biases \autocite{WeidingerEthicalSocialRisks2021}. Researchers have explored methods to guide LLMs to align more closely with ethical norms while preventing them from generating harmful content. For example, OpenAI's work on fine-tuning language models with human feedback has demonstrated that incorporating human preferences into the training process significantly improves the models' alignment with desired behaviors \autocite{OuyangTrainingLanguageModels2022}. Similarly, \textcite{BaiConstitutionalAIHarmlessness2022} explored methods for training models to follow ethical principles through self-improvement without relying on human-labeled data to identify harmful content. However, despite these advancements, challenges remain in ensuring consistency and handling complex ethical dilemmas that require nuanced understanding, making this an active area of ongoing research \autocites{BommasaniOpportunitiesRisksFoundation2022}{WangAligningLargeLanguage2023}{KirkBenefitsRisksBounds2024}.

\subsubsection{Simulating Human Behaviors in Social Contexts} \label{sec:simulating_human_behaviors}

Another stream of research focuses on examining the performance of LLMs in human behavioral experiments or real-life scenarios, comparing their actions to those of humans in various social and economic contexts. For instance, scholars suggest that LLMs can serve as ``computational models of humans,'' simulating human-like behavior in economic games and, at times, demonstrating more cooperative and altruistic behavior than humans \autocites{HortonLargeLanguageModels2023}{MeiTuringTestWhether2024}{JohnsonEvidenceBehaviorConsistent2023}{XieCanLargeLanguage2024}{MageeStructuredLanguageModel2023}. However, LLMs can also be ``too human''---these agents may exhibit ``hyper-accuracy distortion,'' where they simulate human subjects but provide unnaturally accurate responses in classic economic and psychological experiments \autocite{AherUsingLargeLanguage2023}.

Although some scholars propose that LLMs are most useful ``when studying specific topics, when using specific tasks, at specific research stages, and when simulating specific samples'' \autocite[597]{DillionCanAILanguage2023}, this has not deterred researchers from assembling LLM agents into systems that resemble human societies \autocite{GuoLargeLanguageModel2024}. These agents collaboratively interact with each other in various social contexts without specific experimental tasks, such as communicating information \autocite{PerezWhenLLMsPlay2024}, generating novel ideas \autocite{NisiotiCollectiveInnovationGroups2024}, collaborating on software development \autocite{QianCommunicativeAgentsSoftware2023}, and even simulating communal life \autocites{ParkGenerativeAgentsInteractive2023}{LaiEvolvingAICollectives2024}.

Existing studies have demonstrated that LLMs can mimic human behaviors and be guided to align with human values to some extent, but significant challenges remain. Their responses are highly sensitive to prompt phrasing, making it difficult to ensure consistency and to handle complex ethical dilemmas that require nuanced understanding. Moreover, by focusing primarily on LLMs' external behaviors and leaving their internal decision-making processes as a black box, we cannot fully comprehend their actions and confidently deploy them in critical decision-making scenarios. This underscores the necessity for approaches that delve into the inner workings of LLMs rather than merely evaluating their outputs.

\subsection{Framing Research: LLM Agents in Dictator Games} \label{sec:framing_research}

\subsubsection{Two Routes to ``Epistemic Opacity'': Prediction and Explanation}

A notable similarity between these LLM agents and humans is that they are both \textit{epistemically opaque}, which refers to the inherent difficulty in fully understanding or predicting the internal decision-making processes of complex systems \autocite[618]{HumphreysPhilosophicalNoveltyComputer2009}.\footnote{``Epistemic opacity'' can be formally defined as ``a process is epistemically opaque relative to a cognitive agent \textit{X} at time \textit{t} just in case \textit{X} does not know at \textit{t} all of the epistemically relevant elements of the process. A process is essentially epistemically opaque to \textit{X} if and only if it is impossible, given the nature of \textit{X}, for \textit{X} to know all of the epistemically relevant elements of the process'' \autocite[618]{HumphreysPhilosophicalNoveltyComputer2009}.} In humans, this opacity arises from the intricate interplay of cognitive functions, emotions, and subconscious influences that govern behavior. Similarly, LLM agents exhibit epistemic opacity due to the complexity of their neural network architectures and the vastness of their training data, making it challenging to trace how specific inputs lead to particular outputs.

In addressing this epistemic opacity, computer scientists and social scientists have taken different routes \autocite[181]{HofmanIntegratingExplanationPrediction2021}. Computer scientists are more concerned with developing accurate predictive models, whether or not they correspond to causal mechanisms or are even interpretable. The \textit{prediction paradigm} emphasizes the ability to forecast outcomes accurately, often relying on complex models that may be opaque but yield high predictive performance. On the other hand, social scientists have traditionally prioritized interpreting individual and collective human behavior, often invoking causal mechanisms derived from substantive theory and empirical evidence. This \textit{explanation paradigm} values understanding the underlying causes and mechanisms that drive behavior, aiming for interpretability and theoretical insight.

While both paradigms have their own merits---the prediction paradigm excels in accuracy and practical utility, and the explanation paradigm offers deeper understanding and interpretability---relying heavily on prediction is insufficient for understanding the behaviors of LLM agents in complex social contexts. Predictive models may forecast outcomes effectively but often lack transparency and are highly dependent on the datasets they are trained on, which can limit the generalizability of predictions to new or varied contexts. Although significant advancements have been made in explainable AI and its real-world applications \autocites{RibeiroWhyShouldTrust2016}{AmarasingheExplainableMachineLearning2023}{BrandRecentDevelopmentsCausal2023}, the emphasis remains on identifying effective features that contribute to the prediction of specific outcomes. It provides some level of interpretability but falls short of offering insights into how and why certain decisions are made.

From the perspective of social scientists, although individual human behavior is difficult to predict accurately, general patterns and social norms at group level can be systematically studied and interpreted. Empirical social scientists have been analyzing human societies for over a century using methods that consider a wide range of variables, such as demographics, personality traits, and social context. Such evaluation of variables includes understanding the \textit{interactions between these variables} (e.g., interaction terms in regression models), their \textit{partial effects} (e.g., coefficients of variables in regression models), and their \textit{collective impact on outcomes} (e.g., a regression model's goodness of fit). To better understand and anticipate their behavior---especially if we expect LLM agents to be as intelligent and collaborative as humans---we need an approach that integrates social scientists' explanation paradigm, moving beyond the benchmark and validation tests.

\subsubsection{Toward Behavioral Evaluation of LLMs}

New evaluation paradigms are needed---ones that systematically assess these models in realistic and socially complex scenarios. Behavioral experiments, such as simulating economic games, social interactions, and psychological experiments, offer a promising avenue. Evaluating models in settings that mirror human social behaviors enables researchers to explore:

\begin{enumerate}[topsep=0pt,itemsep=0pt,partopsep=0ex,parsep=0ex]
    \item \textit{Decision-Making Processes and Internal Mechanisms}: Examining the underlying factors that influence a model's decisions, allowing for analysis beyond mere input-output patterns to reveal internal dynamics.
    \item \textit{Social Contexts}: Understanding how models navigate ethical dilemmas, fairness considerations, and cooperative settings.
    \item \textit{Alignment with Human Cognitive Processes}: Evaluating whether the models' internal processes and decision-making patterns align with human cognition and behavior.
\end{enumerate}

\subsubsection{LLM Agents in Dictator Games: Sense of Self and Theory of Mind Designs}

In this study, we operationalize the behavioral evaluation of LLM agents by examining their performance in a classic economic experiment: the \textit{dictator game}. Social scientists have widely used this experiment to study prosocial behavior and notions of fairness, which are fundamental social norms in human societies. In a classic dictator game, one participant (the \textit{dictator}) is given a certain amount of money or resources and must decide how much, if any, to share with another participant (the \textit{recipient}), who has no power to influence the decision. Appendix \ref{sec:human_baseline_factors} provides a detailed review of the factors that influence human behavior in this experiment, establishing a ``ground truth'' for our comparative analysis.

\label{rl:external_validity} The dictator game, while elegant in its simplicity for refuting the notion of pure self-interest, is not without its limitations. It represents a somewhat artificial scenario with limited real-world parallels \autocite{LevittWhatLaboratoryExperiments2007}. While some studies show a modest correlation between experiment behaviors and real-life prosocial actions \autocite{WangIncreasingExternalValidity2023}, others find no such link \autocite{GalizziExternalValiditySocial2019}. However, we want to clarify that the major purpose of this study is \textit{not to test the external validity} of the dictator game but to use it as a controlled setting to \textit{compare the behaviors of LLM agents with human baselines}.

\label{rl:llm_dg_studies} Many studies have already begun to explore the behaviors of LLMs in dictator games or similar experiments. Early studies generally found that LLMs often behave like ``typical humans,'' mimicking human behavior in various classic economic games \autocites{HortonLargeLanguageModels2023}{JohnsonEvidenceBehaviorConsistent2023}. For example, \textcite{BrookinsPlayingGamesGPT2023} observed that LLMs exhibit a tendency toward fairness in the dictator game, sometimes even more so than human participants \autocite{MeiTuringTestWhether2024}. LLMs agents also demonstrate reasoning abilities in strategic settings \autocite{SreedharSimulatingHumanStrategic2024}. However, their behavior is highly sensitive to the contents of prompts and varies significantly across different models of varying sizes \autocites{ChanScalableEvaluationCooperativeness2023}{FanCanLargeLanguage2024}. Further research indicates that while LLMs can replicate many psychological experiments, they often produce larger effect sizes than human studies and show lower replication rates for socially sensitive topics \autocite{CuiCanLargeLanguage2025}. Specifically in dictator games, even advanced models like GPT-4o fail to accurately predict human behavior, consistently underestimating self-interest and overestimating altruism, a phenomenon described as an ``optimistic bias'' \autocite{CapraroPubliclyAvailableBenchmark2025}.

Building upon the fruitful scholarship, we aim to understand \textit{what causes the variations in LLM agents' behavior in dictator games?} We address this question by framing our research design around two primary psychological perspectives: \textit{Sense of Self (SoS)} and \textit{Theory of Mind (ToM)}.

From the SoS perspective, we explore how different persona settings of LLM agents influence their decision-making processes. Sense of Self refers to an individual's perception and awareness of their own identity, including traits, beliefs, and social roles. This self-concept affects how individuals interpret situations and make decisions \autocite{MarkusDynamicSelfConceptSocial1987}. In the context of LLMs, we simulate this by assigning different personas to the agents, allowing us to examine whether and how these self-concepts affect their choices in the dictator game.

From the ToM perspective, we investigate whether LLM agents can model the behavior of humans with different backgrounds. Theory of Mind is the ability to attribute mental states---such as beliefs, intents, desires, and knowledge---to oneself and others, understanding that others have perspectives different from one's own and enabling the predictions about the behavior of others \autocites{PremackDoesChimpanzeeHave1978}{ApperlyWhatTheoryMind2012}. This cognitive ability is crucial for social interactions and empathy. By assessing the LLMs' capacity to anticipate human behavior based on contextual information, we evaluate their ability to emulate ToM in decision-making scenarios and extend existing studies \autocite{StrachanTestingTheoryMind2024}.

By comparing the performance of LLM agents in dictator games across these two psychological perspectives and with human baselines, we aim to understand the decision-making processes of LLM agents and identify the factors that influence their prosocial behaviors. This approach not only helps us unpack the internal mechanisms driving LLM behavior but also contributes to the broader understanding of how artificial intelligence can replicate complex---not only the behaviors of humans, but also the internal psychological processes of humans.

\clearpage
\section{Methods}

\subsection{Experiment Design}

\begin{figure}[htbp]
    \centering
    \caption{\textsc{Experiment Design: LLM Agent in Dictator Game}}
    \label{fig:expr_design}
    \includegraphics[width=.9\textwidth]{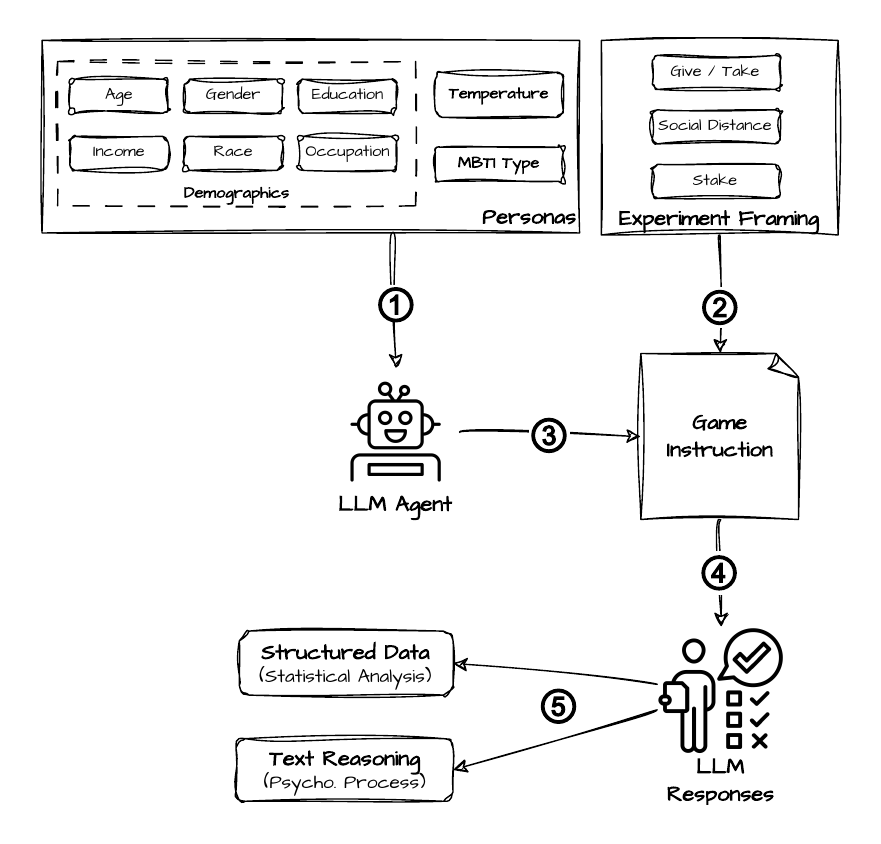}
    \vspace{0.5em} 
    \begin{minipage}{.9\textwidth}
        \footnotesize\textit{Note}: Numbers in circles indicate the order of steps. See Appendix \ref{sec:human_baseline_factors} and Section \ref{sec:regression_analysis} for detailed descriptions of the variables and experimental settings.
    \end{minipage}
\end{figure}

We selected the 10 most popular open-source LLM models in varied sizes from four families (i.e., Llama3.1, Gemma2, Qwen2.5, and Phi3), along with GPT4o (Appendix \ref{sec:llm_selection}), to participate in the experiment as Figure \ref{fig:expr_design} illustrates.\footnote{Additionally, we included the most advanced open-source reasoning model available as of May 2025, \texttt{DeepSeek-R1-70B}, as a robustness check. This model employs a reasoning-chain and mixture-of-experts architecture in its training, enhancing efficiency and performance across multiple benchmarks\autocite{DeepSeek-AIDeepSeekR1IncentivizingReasoning2025,DaiDeepSeekMoEUltimateExpert2024}. However, despite its advanced reasoning capabilities, it does not demonstrate more human-like behaviors, as evidenced by Column 11 in Table \ref{tab:alignment_self}.} Each experimental trial follows the steps below:

\begin{enumerate}[topsep=0pt,itemsep=0pt,partopsep=0ex,parsep=0ex]
    \item \textit{Setting Persona of LLM Agent}: Randomly select a combination of demographic variables, LLM temperature values, and personality traits to define the persona of an LLM agent. Listings \ref{prom:agent_setting_instruction_self} and \ref{prom:agent_setting_instruction_tom} in the appendix are used to set the personas of LLM agents based on the SoS and ToM perspectives, respectively.
    \item \textit{Framing Experiment Instruction}: Construct the experiment instructions (Section \ref{sec:framing}) by randomly selecting options for social distance and Give vs. Take framing, and by setting a random stake amount (elaborated in the following section). We prepared four game instructions by psychological perspectives (i.e., SoS and ToM) and the framing of games (i.e., Give and Take). The instructions are presented to the LLM agent using Listings \ref{prom:game_instruction_self_give}---\ref{prom:game_instruction_tom_take} in the appendix.
    \item \textit{Game-Play and Collecting LLM Responses}: Present the experiment instruction to the LLM agent and collect its responses. The collected responses consist of two parts: (1) structured data in JSON format, including variables such as the agent's age, education level, and the amount of money transferred; and (2) textual data, which captures the agent's reasoning behind its decisions (see Listings \ref{prom:example_reasoning1}---\ref{prom:example_reasoning3} in the appendix for three examples).
\end{enumerate}

Tables \ref{tab:descriptive_age_self}--\ref{tab:amount_transfer_tom} in the appendix present the descriptive statistics of key variables and experimental results of each LLM model. Except for models with a small number of logically correct trials (e.g., \texttt{phi3\_3.8b} and \texttt{qwen2.5\_7b}), the distributions of most variables across different models are well-balanced. This ensures that the results are not biased because of the distribution of variables across models.

\subsection{Factors Influencing LLM Generosity} \label{sec:regression_analysis}

Based on the review of human empirical studies on dictator games (Appendix \ref{sec:human_baseline_factors}), we identified key predictors from three aspects: LLM personas, experiment framing, and psychological process. 

\subsubsection{LLM Personas} \label{sec:personas}

\textbf{Demographics.} To generate demographic profiles for the LLM agents, we used options from two large-scale U.S. public surveys: the General Social Survey (GSS) and the American Community Survey (ACS). The GSS, widely recognized in social science research, includes both attitudinal data (such as happiness and views on marriage and social issues) and background information (such as marital status, race, and education). It has been supporting a wide range of research topics, such as income inequality, educational attainment, immigration, and religious beliefs \autocite{MarsdenTrackingUSSocial2020}. The ACS, conducted annually by the U.S. Census Bureau, provides comprehensive data on economic, social, housing, and demographic characteristics of the U.S. population and is an essential resource for policymakers \autocite{NationalResearchCouncilUsingAmericanCommunity2007}.

Given their extensive use in academia and established reliability, we selected nine variables from these surveys to construct demographic pools for developing the personas of LLM agents . These variables include age (continuous: between 20 and 60), gender (binary: male or female), education (ordinal: less than high school, high school, and bachelor's degree or higher), marital status (binary: currently married or unmarried), race (categorical: 15 racial groups), household income (ordinal: 10 categories), Hispanic status (binary: Hispanic or Latino vs. not Hispanic or Latino), occupation (categorical: 5 occupations), and industry (categorical: 13 industries). In each trial, we randomly generated a demographic profile for an agent using these nine variables. It enables us to explore how the demographic settings of LLM agents, in combination with other traits and experimental contexts, influence their decisions in dictator games.

\textbf{Temperature.} This is a unique setting that defines the randomness of an LLM's output. A lower temperature (close to 0) makes a model's responses more deterministic and focused on the most likely outcomes. Conversely, a higher temperature increases the randomness, allowing for more diverse and creative outputs by giving less probable words a greater chance of being selected. Although the temperature setting is theoretically meaningful, empirical studies have found that its impact is minimal in various real-world tasks \autocites{PatelExploringTemperatureEffects2024}{PeeperkornTemperatureCreativityParameter2024}{RenzeEffectSamplingTemperature2024}. In this study, we randomly assign this hyperparameter a value between 0 and 1.00 for each trial to examine how variations in temperature affect agents' decisions in conjunction with their other traits.

\textbf{MBTI Personality Types.} Existing studies on prosocial behaviors commonly use the Big Five model to measure personality traits, while the MBTI is more popular in human resource studies. Correlation analyses have shown strong relationships between the two psychological scales, such as Big Five Extraversion correlating with MBTI Extraversion-Introversion, and Openness to Experience correlating with Sensing-Intuition \autocite{FurnhamBigFiveBig1996}.

We adopt MBTI in this study for several reasons, particularly its practical advantages in computational studies \autocite[93]{CelliBigFiveBetter2018}. The Big Five model defines personality along five scales: Openness to Experience, Conscientiousness, Extraversion, Agreeableness, and Neuroticism. In contrast, the MBTI categorizes personality into four binary dimensions---Extraversion/Introversion, Sensing/Intuition, Thinking/Feeling, and Judging/Perceiving---resulting in 16 distinct personality types. Since MBTI types are represented as simple 4-letter codes (e.g., INTJ), it is much easier to collect gold-standard labeled data (i.e., training datasets) for developing machine learning classifiers.

In this study, we randomly select one of the 16 MBTI types in each trial to define the personality of the LLM agent. This approach allows us to explore how different personality types, as defined by MBTI, influence the prosocial behaviors of LLM agents in conjunction with other personal traits and experimental settings.

\subsubsection{Experiment Framing} \label{sec:framing}

\label{rl:social_distance}
\textbf{Social Distance.} We construct this variable based on ``the degree of reciprocity that subjects believe exists within a social interaction'' \autocite[654]{HoffmanSocialDistanceOtherRegarding1996}. \label{sec:framing_social_distance} Our study includes three levels of social distance, with the exact instruction texts shown below; these imply different levels of anonymity in line with dictator-game practice:

\begin{itemize}[topsep=0pt,itemsep=0pt,partopsep=0ex,parsep=0ex]
    \item \textit{Stranger}: ``You and the other person are strangers, and you two will not interact after this game.''---total anonymity.
    \item \textit{Stranger Meet Afterward}: ``You and the other person are strangers, but you two will meet each other after this game.''---partial anonymity.
    \item \textit{Friends}: ``You and the other person are friends.''---no anonymity.
\end{itemize}

This design jointly varies social distance and the presence of a minimal social cue (anticipated meeting) alongside the implied anonymity level, consistent with evidence that contextual cues and anonymity meaningfully shape transfers in dictator games \autocites{RigdonMinimalSocialCues2009}{FranzenAnonymityDictatorGame2012}.

\textbf{Give vs. Take.} To examine the effects of ``Give'' vs. ``Take'' framing on the agents' decisions, we designed the game instructions based on \textcite{CappelenGiveTakeDictator2013}. In a ``Give'' game, agents are informed that both they and the recipients have the same initial amount of money. However, the agents also receive an additional amount (i.e., the Stake), which the recipients do not. The dictator can transfer any amount, from 0 up to the total amount of their additional money, to the recipients. In a ``Take'' game, the instructions follow the same structure, but the difference is that agents can transfer a negative amount, meaning they can take money from the recipients.

\textbf{Stake.} To ensure comparability with most existing studies, we randomly generate an integer between 10 and 100 USD as the initial amounts of money (i.e., the ``initial endowment'' commonly referred to in existing studies) and the additional amounts of money (i.e., the ``stake'' commonly referred to in existing studies) as specified in the game instructions.

\subsubsection{Psychological Processes} \label{sec:psychological_processes}

The LLM agents were instructed to explain their decisions, providing unstructured text responses that are useful for understanding their psychological processes.\footnote{\label{rl:llm_reasoning}For the model's stated reasoning, this output should be interpreted with caution. Current LLMs are optimized to produce plausible and convincing text, which may lead to post-hoc rationalizations that do not faithfully reflect the model's actual computational process \autocite{TurpinLanguageModelsDont2023}.} To analyze these responses, we used the Linguistic Inquiry and Word Count \autocite[LIWC;][]{TausczikPsychologicalMeaningWords2010}, a widely recognized text analysis instrument in psychology. LIWC helps to infer individuals' psychological states based on language use by categorizing words into various psychological dimensions, such as cognitive, emotional, and social processes. It allowed us to explore the psychological states underlying the agents' decisions in dictator games.

We specifically focused on LIWC categories relevant to compassion and empathy, which are fundamental in shaping prosocial behaviors \autocite{YadenCharacterizingEmpathyCompassion2024}. The compassion-related categories include Positive Emotion (e.g., love, good, happy), Social Processes (e.g., you, your, love, they), Religion (e.g., God, hell, pray), Affiliation (e.g., our, friends, family), Certainty (e.g., all, never, always), Family (e.g., baby, dad, mom), Drives (e.g., up, get, good), and Affect (e.g., love, happy, great). The empathy-related categories include First-Person Singular (e.g., I, my, me), Focus on the Present (e.g., is, be, are), Personal Pronouns (e.g., I, you, me), Sadness (e.g., miss, lost, sorry), Discrepancy (e.g., should, would, could), Verbs (e.g., is, have, was), Adverbs (e.g., so, just, about), Cognitive Processes (e.g., cause, know, ought), Pronouns (e.g., I, them, her), and Affective Processes (e.g., happy, cried, abandon).

\subsection{Empirical Analysis}

To evaluate how different personas and experimental contexts influence the behavior of LLM agents, we conducted regression analyses for each model. The dependent variable was the proportion of the stake transferred by the agent. The independent variables included persona attributes (e.g., age, gender, education, and MBTI type),\label{rl:demographic_variables}\footnote{Most dictator game studies with human participants treat demographic variables primarily as controls, as the main focus is often on isolating the causal effects of experimental manipulations. In contrast, a central research question of our study is to investigate whether LLMs can effectively adopt and act upon specific ``personas.'' Therefore, understanding the influence of these persona attributes on the LLM's behavior is a primary research interest, not secondary.} experimental settings (social distance, Give vs. Take framing, and stake amount), and psychological processes (LIWC category scores). We also included other demographic variables, such as race, occupation, and industry, as controls to account for potential confounding effects.

Furthermore, we compared the regression coefficients with the expected results from human studies (Appendix \ref{sec:human_baseline_factors}) to evaluate the alignment between LLM agents and human participants. This comparison helps us understand the extent to which LLM agents' decision-making processes and internal mechanisms align with those of humans.

\clearpage
\section{Results}

\subsection{Model Performance}

\subsubsection{Instruction Following and Math Reasoning}

Table \ref{tab:model_performance_self} summarizes each model's performance on instruction following and math reasoning. Instruction following is measured by the number of responses returned in valid JSON, because agents were instructed to output JSON. Math reasoning is measured by the number of logically correct trials, i.e., cases where payoffs are computed correctly under the stated framing. For example, in a ``Take'' game where both players initially receive \$100 and the dictator also receives a \$100 stake, a transfer of -\$20 yields a recipient payoff of \$80 (= 100 - 20) and a dictator payoff of \$220 (= 100 + 100 + 20). \label{rl:math_question} We treat math reasoning as a manipulation check: it is not the study's primary focus, but it is necessary to verify that models can perform the basic arithmetic reliably, and only logically correct trials are included in the subsequent analyses.

The results in Table \ref{tab:model_performance_self} show that while all models exhibit a strong ability to follow instructions,\footnote{GPT4o includes a setting that enforces output in JSON format, but we did not use this feature to maintain comparability with other open-source models.} their math reasoning capabilities vary considerably. Surprisingly, \texttt{Llama3.1-70B} achieves the highest percentage of logically correct trials (96.36\%) among all the models, surpassing even industry frontier model, \texttt{GPT4o-2024-08-06}, and the significantly larger \texttt{Llama3.1-405B} in the Llama family. The \texttt{Qwen2.5-7B} model demonstrates the lowest performance in math reasoning, with only 5.37\% of logically correct trials. \label{rl:model_size} In general, while model size plays an important role in performance, it is not the sole determining factor---smaller models can sometimes outperform larger ones. There appears to be an optimal size that balances performance and computational efficiency \autocite{HoffmannTrainingComputeOptimalLarge2022}.

\begin{table}[!h]
    \caption{\textsc{Model Performance: Instruction following and math reasoning}} \label{tab:model_performance_self}
    \small
    \begin{threeparttable}
        \begin{tabular}{>{\raggedleft\arraybackslash}m{1cm}m{3cm}>{\raggedleft\arraybackslash}m{2cm}>{\raggedleft\arraybackslash}m{2cm}>{\raggedleft\arraybackslash}m{2cm}>{\raggedleft\arraybackslash}m{2cm}}
            \hline\hline
               & {Model\_Size}     & {\#Simulation Trials} & {\#Correct JSON Format} & {\#Logically Correct Trials} & {\%Logically Correct Trials} \\
            \hline
            1  & llama3.1\_70b     & 10,000                & 9,997                   & 9,633                        & 96.36                        \\
            2  & gpt4o\_2024-08-06 & 10,000                & 10,000                  & 9,561                        & 95.61                        \\
            3  & llama3.1\_405b    & 10,000                & 9,977                   & 8,997                        & 90.18                        \\
            4  & gemma2\_27b       & 10,000                & 9,996                   & 8,271                        & 82.74                        \\
            5  & qwen2.5\_72b      & 10,000                & 10,000                  & 5,442                        & 54.42                        \\
            6  & gemma2\_9b        & 10,000                & 9,736                   & 4,582                        & 47.06                        \\
            7  & llama3.1\_8b      & 10,000                & 9,944                   & 4,020                        & 40.43                        \\
            8  & phi3\_14b         & 10,000                & 9,808                   & 2,980                        & 30.38                        \\
            9  & phi3\_3.8b        & 10,000                & 9,820                   & 773                          & 7.87                         \\
            10 & qwen2.5\_7b       & 10,000                & 9,956                   & 535                          & 5.37                         \\
            \hline\hline
        \end{tabular}
        \begin{tablenotes}[para,flushleft]
            \footnotesize\textit{Note}: ``\#Correct JSON Format'' indicates the number of responses in correct JSON format, suggesting a model's ability of instruction following. ``\#Logically Correct Trials'' and ``\%Logically Correct Trials'' indicate the number and corresponding percentage of responses that are logically correct, suggesting a model's ability of math reasoning. Results of the Theory of Mind trials are in Appendix Table \ref{tab:model_performance_tom}.
        \end{tablenotes}
    \end{threeparttable}
\end{table}

\subsubsection{Giving Rate}

\begin{sidewaysfigure}[!htbp]
    \centering
    \caption{\textsc{Giving rate by model family and size (SoS)}}
    \label{fig:giving_rate_self}
    \includegraphics[width=\textwidth]{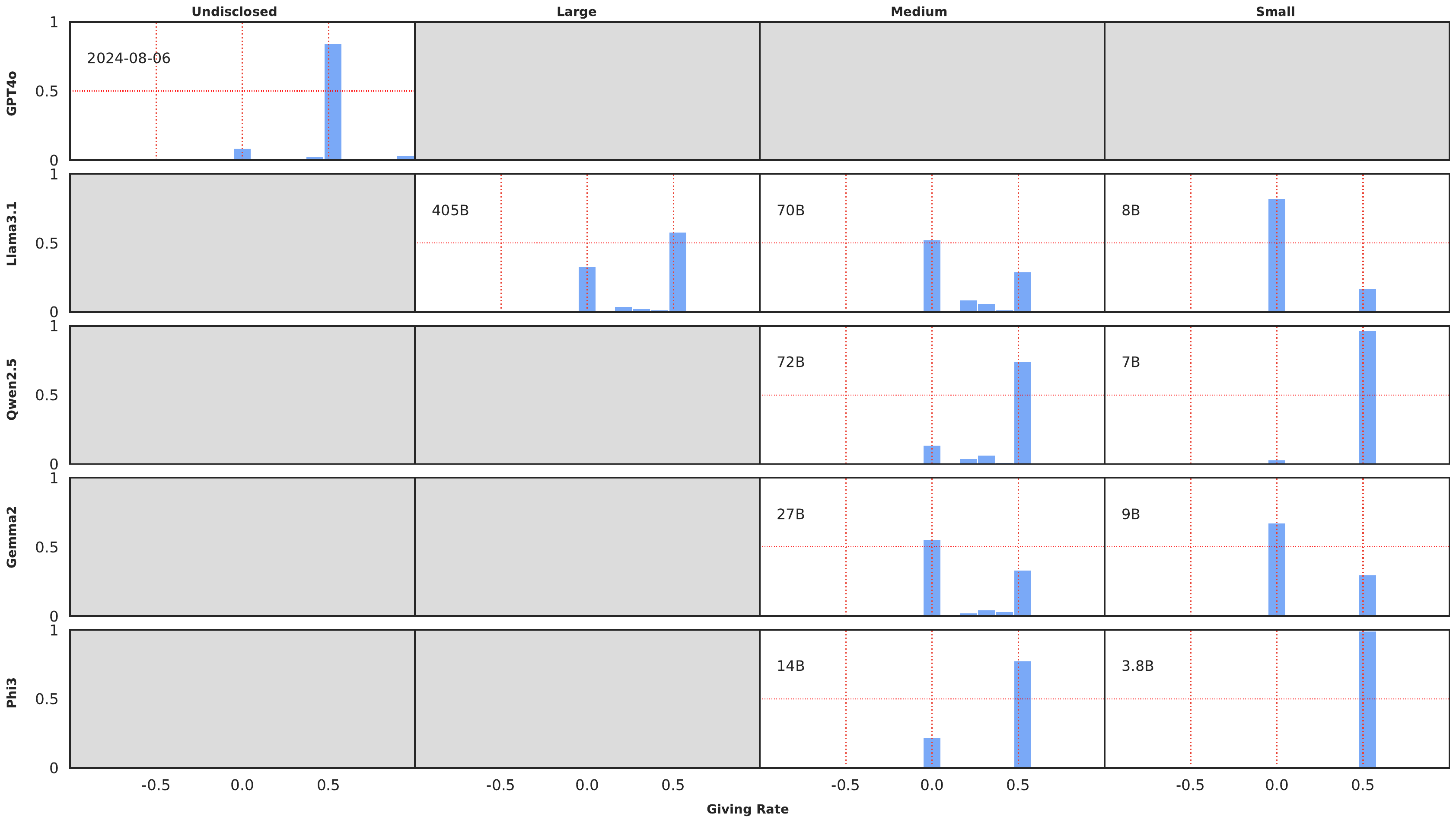}
    \vspace{0.5em} 
    \begin{minipage}{\textwidth}
        \footnotesize\textit{Note}: Vertical red dashed lines indicate giving rates at -0.5, 0, and 0.5, respectively; horizontal red dashed lines indicate 50\% of total observations. The giving rate is calculated as the percentage of the amount transferred by the dictator to the recipient out of the total stake. Results of the Theory of Mind trials are in Appendix Figure \ref{fig:giving_rate_tom}.
    \end{minipage}
\end{sidewaysfigure}

Figure \ref{fig:giving_rate_self} shows the giving rates of each LLM model by family and size. The giving rate is calculated as the percentage of the amount transferred by the dictator to the recipient out of the total stake. As the figure presents, the decision space (i.e., the distribution of giving rates) for most of these models is bimodal, with choices concentrated at 0 (i.e., giving nothing) and 0.5 (giving half), showing the problem of ``hyper-consistent responses'' or ``uniformity'' \autocites[19]{KozlowskiSimulatingSubjectsPromise2024}{BisbeeSyntheticReplacementsHuman2024}. This pattern differs significantly from that observed in human behavior, where the distribution of giving rates is continuous and clustered around 0 (36.11\%), 0.5 (16.74\%), and 1 (i.e., giving all; 5.44\%) \autocite[589]{EngelDictatorGamesMeta2011}. The 70B model of the Llama family exhibit the most continuous distribution of giving rates, although they still deviate from human behavior. Additionally, the decision space varies significantly even within the same model family, with no clear pattern from smaller to larger models.

Overall, LLM agents are unable to capture the continuous distribution of human behavior and lack variation in decision-making, which consequently increases the certainty of their decisions. Conversely, there is a lack of consistency within the same model family, increasing the uncertainty of predicting LLM behaviors. These paradoxical results present practical implications on LLM evaluation and alignment with human behavior and will be discussed later (Section \ref{sec:determinism}).

\subsection{Predicting the Behavior of LLM Agents: Sense of Self Trials}

Given the SoS and ToM trials follow the same experimental and analytical structure, we present the results of the SoS trials in this section, with the ToM trial results provided in Appendix \ref{sec:tom_results}. In the main text, we focus on comparing the outcomes of the two designs.

\subsubsection{Personas}

\begin{figure}[!htbp]
    \centering
    \caption{\textsc{Predicting generosity: Demographics and LLM temperature (SoS)}}
    \label{fig:demographics_self}
    \includegraphics[width=1\textwidth]{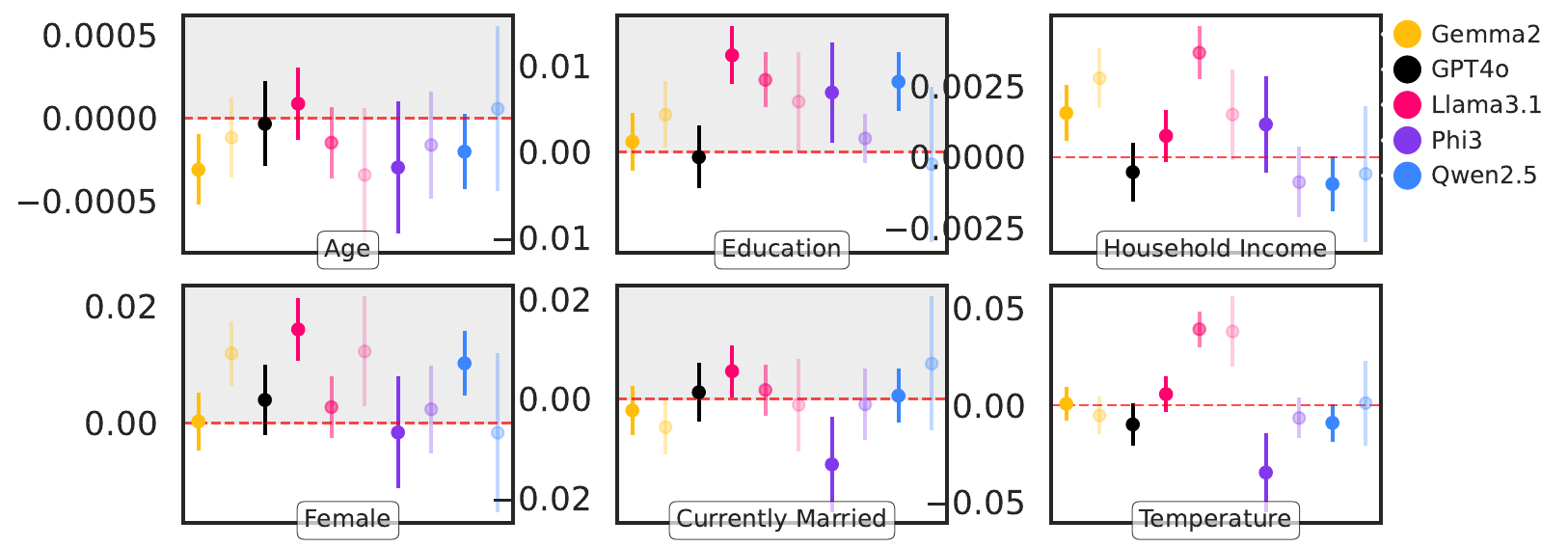}
    \begin{minipage}[t]{\linewidth}
        \footnotesize\textit{Note}: The coefficients (showing 95\% confidence intervals) are from a linear regression model using the proportion of stake transferred in the dictator game as the dependent variable. Deep colors represent larger models, and light colors represent smaller models within the same LLM family. The shaded areas indicate expected directions of impact based on human studies (Appendix \ref{sec:human_baseline_factors}). Results of the Theory of Mind trials are in Appendix Figure \ref{fig:demographics_tom}.
    \end{minipage}
\end{figure}

\textbf{Demographics.} Figure \ref{fig:demographics_self} displays the coefficients of the demographic variables and LLM temperature in predicting generosity. Few of these models exhibit behavior consistent with human studies. Among them, \texttt{Llama3.1-70B} and \texttt{Llama3.1-405B} are the most human-like, showing performance consistent with humans on \textit{Education}, \textit{Household Income}, and \textit{Female}. The industry frontier model, \texttt{GPT4o-2024-08-06}, does not align with human behavior on any of these demographic variables. Whether this is surprising or not can depend on how we posit the debiasing efforts in developing the larger models---debiasing in LLMs involves reducing stereotypes and biases from the training data by adjusting data sampling or applying fairness constraints \autocite{MeadeEmpiricalSurveyEffectiveness2022}. These efforts aim to make models more neutral, though they can result in deviations from typical human patterns.

Figure \ref{fig:demographics_self} also shows substantial variations and inconsistencies in the coefficients at different levels. First, the coefficients of the same demographic variable differ significantly across different model families. For example, for \textit{Household Income}, models from Gemma2 and Llama families show positive impact, while Phi3 and Qwen2.5 models show the opposite. Second, the coefficients of the same demographic variable differ significantly even within the same LLM family. For instance, the coefficients for \textit{Female} differ substantially within the Llama3.1 family---the 405B model shows a positive effect on the money transferred, the 70B model shows no significance, while the 7B model shows a positive effect again. Third, for agents driven by the same LLM model, their behaviors are not deterministic and can vary significantly. For example, \texttt{Phi3-14B} exhibit large variations in the coefficients for all demographic variables.

\textbf{LLM Temperature.} For the coefficients of \textit{Temperature}, as shown in Figure \ref{fig:demographics_self}, the differences across the models are mixed, with some models demonstrating opposite effects. The coefficients for Llama models indicate a significant positive relation between the value of \textit{Temperature} and the amount of money transferred, whereas the coefficient of GPT4o is negative. These contrasting effects suggest that the influence of temperature settings on model behavior is variable and model-dependent. Although the actual effect may be limited due to the narrow range of possible \textit{Temperature} values (i.e., between 0 and 1), the inconsistency across models raises concerns about the reliability and interpretability of LLM agents.

\begin{figure}[!htbp]
    \centering
    \caption{\textsc{Predicting generosity: Myers–Briggs Type Indicator (SoS)}}
    \label{fig:mbti_self}
    \includegraphics[width=0.8\textwidth]{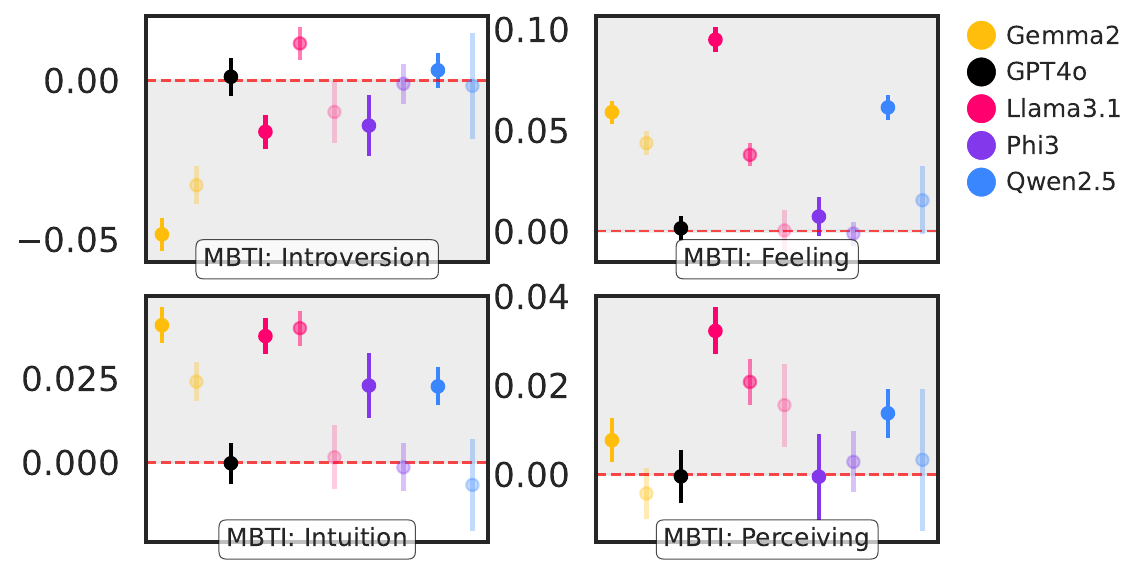}
    \begin{minipage}[t]{0.8\linewidth}
        \footnotesize\textit{Note}: The coefficients (showing 95\% confidence intervals) are from a linear regression model using the proportion of stake transferred in the dictator game as the dependent variable. Deep colors represent larger models, and light colors represent smaller models within the same LLM family. The shaded areas indicate expected directions of impact based on human studies (Appendix \ref{sec:human_baseline_factors}). Results of the Theory of Mind trials are in Appendix Figure \ref{fig:mbti_tom}.
    \end{minipage}
\end{figure}

\textbf{MBTI Personality Types.} Figure \ref{fig:mbti_self} illustrates the relationships between MBTI personality types and the amount of money transferred in dictator games. The \texttt{Gemma2-27B} and \texttt{Llama3.1-405B} models exhibit the most human-like behaviors, aligning closely with human studies. Specifically, agents driven by the two models with MBTI types Extraversion (E), Intuition (N), Feeling (F), and Perceiving (P) tend to be more generous. In contrast, the other models show insignificance or inconsistent patterns that do not match human studies. For instance, the \texttt{Llama3.1-70B} model shows a positive relationship between Introversion (I) and the amount of money transferred, which contradicts human findings. The industry frontier model, \texttt{GPT4o-2024-08-06}, shows no significance on all MBTI types. These inconsistencies suggest that, from the perspective of personality type, the alignment of LLM agents with human behavior in dictator games varies significantly and is highly model-dependent.

\begin{figure}[!htbp]
    \centering
    \caption{\textsc{Predicting generosity: Framing of experiment (SoS)}}
    \label{fig:framing_self}
    \includegraphics[width=0.8\textwidth]{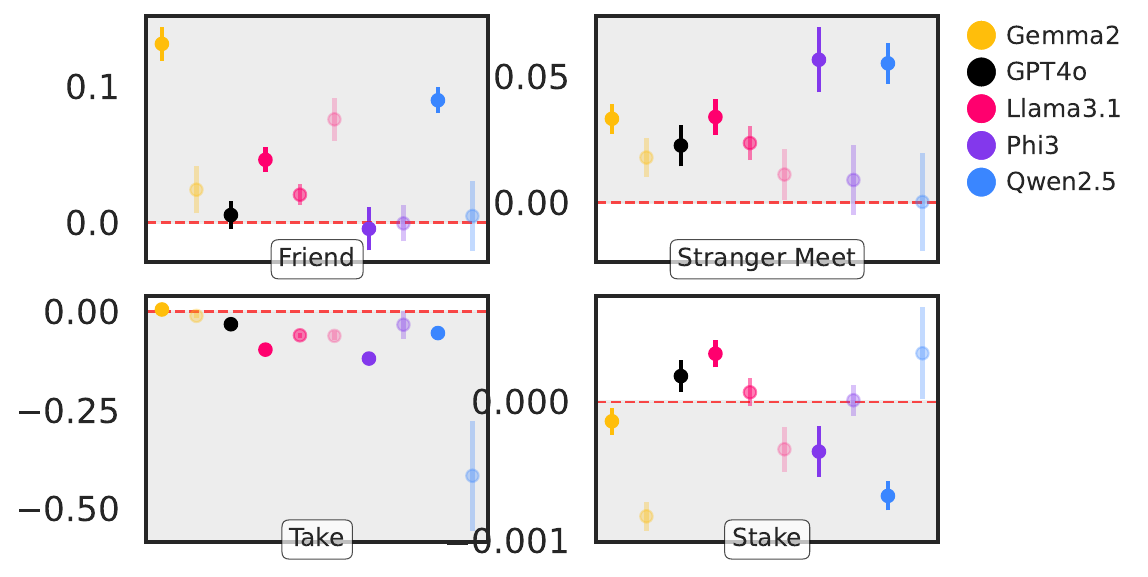}
    \begin{minipage}[t]{\linewidth}
        \footnotesize\textit{Note}: The coefficients (showing 95\% confidence intervals) are from a linear regression model using the proportion of stake transferred in the dictator game as the dependent variable. Deep colors represent larger models, and light colors represent smaller models within the same LLM family. The shaded areas indicate expected directions of impact based on human studies (Appendix \ref{sec:human_baseline_factors}). The ``Stranger'' framing is the reference group for ``Friend'' and ``Stranger Meet.'' The ``Give'' framing is the reference group for ``Take.'' Results of the Theory of Mind trials are in Appendix Figure \ref{fig:framing_tom}.
    \end{minipage}
\end{figure}

\textbf{Experiment Framing.} Figure \ref{fig:framing_self} shows the relationships between the proportion of the stake transferred and various experimental framings. For \textit{Social Distance}, most models behave as expected based on human studies---they tend to give more to known recipients (\textit{Friend}) and recipients they will meet afterward (\textit{Stranger Meet}) than to strangers (\textit{Stranger}). The ``Take'' framing consistently reduces the proportion transferred across most models, closely aligning with human studies. However, the results of \textit{Stake} are mixed, with some models showing a positive relationship and others showing the opposite. These mixed results even occur within the same model family, such as Llama3.1 and Qwen2.5.

\begin{sidewaysfigure}[htbp]
    \centering
    \caption{\textsc{Predicting generosity: Psychological process (SoS)}}
    \label{fig:liwc_self}
    \begin{subfigure}{1\textwidth}
        \centering
        \caption{LIWC Categories Effectively Predicting Compassion Controlling for Empathy}
        \includegraphics[width=1\textwidth]{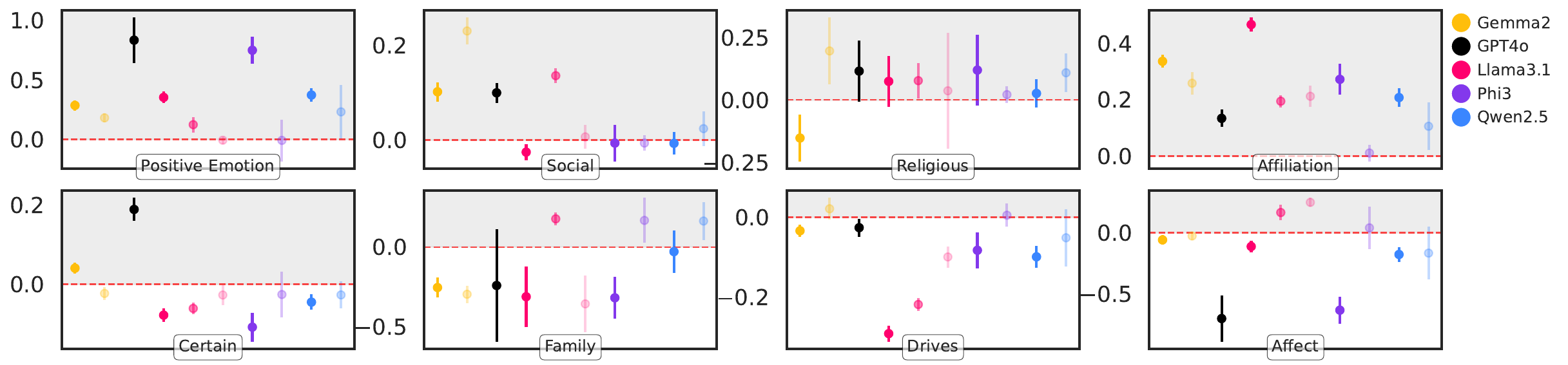}
        \label{fig:compassion}
    \end{subfigure}
    \begin{subfigure}{1\textwidth}
        \centering
        \caption{LIWC Categories Effectively Predicting Empathy Controlling for Compassion}
        \includegraphics[width=1\textwidth]{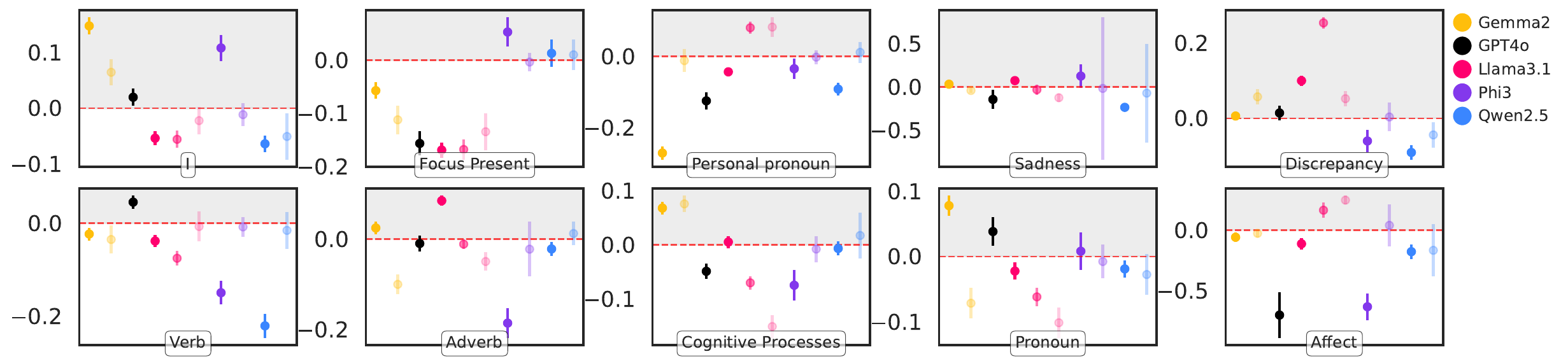}
        \label{fig:empathy}
    \end{subfigure}
    \begin{minipage}[t]{\linewidth}
        \footnotesize\textit{Note}: The coefficients (showing 95\% confidence intervals) are from a linear regression model using the proportion of stake transferred in the dictator game as the dependent variable. Deep colors represent larger models, and light colors represent smaller models within the same LLM family. The shaded areas indicate expected directions of impact based on human studies (Appendix \ref{sec:human_baseline_factors}). LIWC categories are selected for analysis according to \textcite{YadenCharacterizingEmpathyCompassion2024}. ``She/He'' and ``Male'' categories for Compassion are excluded due to limited number of observations. LIWC = Linguistic Inquiry and Word Count \autocite{TausczikPsychologicalMeaningWords2010}. Results of the Theory of Mind trials are in Appendix Figure \ref{fig:liwc_tom}.
    \end{minipage}
\end{sidewaysfigure}

\textbf{Psychological Processes.} Figure \ref{fig:liwc_self} displays the coefficients of LIWC categories in predicting the proportion of money transferred. These categories were chosen to represent the psychological processes of compassion and empathy according to \textcite{YadenCharacterizingEmpathyCompassion2024}. To align with human behavior, all coefficients should be positive. However, the results reveal that all LLM agents display inconsistent patterns. For example, the industry frontier model, \texttt{GPT4o-2024-08-06}, swings between positive and negative coefficients for different LIWC categories, reflecting inconsistencies in the representation of compassion and empathy. The same inconsistency is also observed with the largest and presumably most capable open-source model, \texttt{Llama3.1-405B}. These findings suggest that LLM agents may not fully capture the psychological processes underlying the prosocial behaviors of humans, with their alignment to human behavior being highly variable and model-dependent.

\subsection{Summarizing Sense of Self and Theory of Mind Results}

\begin{sidewaystable}[htbp]
    \caption{\textsc{LLM Agent's Alignment with Humans in Dictator Games (Sense of Self)}} \label{tab:alignment_self}
    \begin{threeparttable}
        \begin{tabular}{rrcccccccccccc}
            \hline\hline
               &                                             & (1)    & (2)    & (3)    & (4)    & (5)    & (6)    & (7)    & (8)   & (9)    & (10)   & (11)   & Total \cmark \\
               &                                             & G27B   & G9B    & GPT4o  & L405B  & L70B   & L8B    & P14B   & P3.8B & Q72B   & Q7B    & R1-70B & (by row)     \\
            \hline
               & \multicolumn{1}{l}{\textit{Demographics}}   &        &        &        &        &        &        &        &       &        &        &        &              \\
            1  & Age                                         & \xmark & n.s.   & n.s.   & n.s.   & n.s.   & n.s.   & n.s.   & n.s.  & n.s.   & n.s.   & \xmark & 0            \\
            2  & Education                                   & n.s.   & \cmark & n.s.   & \cmark & \cmark & \cmark & \cmark & n.s.  & \cmark & n.s.   & \cmark & 7            \\
            3  & H. Income                                   & pos.   & pos.   & n.s.   & n.s.   & pos.   & pos.   & n.s.   & n.s.  & neg.   & n.s.   & pos.   & --           \\
            4  & Female                                      & n.s.   & \cmark & n.s.   & \cmark & n.s.   & \cmark & n.s.   & n.s.  & \cmark & n.s.   & \cmark & 5            \\
            5  & Married                                     & n.s.   & \xmark & n.s.   & \cmark & n.s.   & n.s.   & \xmark & n.s.  & n.s.   & n.s.   & \cmark & 2            \\
            6  & Temperature                                 & n.s.   & n.s.   & n.s.   & n.s.   & pos.   & pos.   & neg.   & n.s.  & n.s.   & n.s.   & neg.   & --           \\
               & \multicolumn{1}{l}{\textit{MBTI}}           &        &        &        &        &        &        &        &       &        &        &        &              \\
            7  & Introversion                                & \cmark & \cmark & n.s.   & \cmark & \xmark & \cmark & \cmark & n.s.  & n.s.   & n.s.   & \xmark & 5            \\
            8  & Feeling                                     & \cmark & \cmark & n.s.   & \cmark & \cmark & n.s.   & n.s.   & n.s.  & \cmark & n.s.   & \cmark & 6            \\
            9  & Intuition                                   & \cmark & \cmark & n.s.   & \cmark & \cmark & n.s.   & \cmark & n.s.  & \cmark & n.s.   & \cmark & 7            \\
            10 & Perceiving                                  & \cmark & n.s.   & n.s.   & \cmark & \cmark & \cmark & n.s.   & n.s.  & \cmark & n.s.   & \cmark & 6            \\
               & \multicolumn{1}{l}{\textit{Exper. Framing}} &        &        &        &        &        &        &        &       &        &        &        &              \\
            11 & Friend                                      & \cmark & \cmark & n.s.   & \cmark & \cmark & \cmark & n.s.   & n.s.  & \cmark & n.s.   & \cmark & 7            \\
            12 & Stranger Meet                               & \cmark & \cmark & \cmark & \cmark & \cmark & \cmark & \cmark & n.s.  & \cmark & n.s.   & \cmark & 9            \\
            13 & Take                                        & n.s.   & n.s.   & \cmark & \cmark & \cmark & \cmark & \cmark & n.s.  & \cmark & \cmark & \cmark & 8            \\
            14 & Stake                                       & \cmark & \cmark & \xmark & \xmark & n.s.   & \cmark & \cmark & n.s.  & \cmark & \xmark & \cmark & 6            \\
            \hline
               & Total \cmark                                & 7      & 8      & 2      & 10     & 7      & 8      & 6      & 0     & 9      & 1      & 10     & 68           \\
            \hline\hline
        \end{tabular}
        \begin{tablenotes}[para,flushleft]
            \footnotesize\textit{Note}: \cmark = Aligning with human studies; \xmark = Not aligning with human studies; n.s. = Not significant; pos. = Positive; neg. = Negative. ``--'' indicates the lack of consensus from human studies, showing directions of coefficients but not alignments for these variables. The expected directions of impact based on human studies are reviewed in Appendix \ref{sec:human_baseline_factors}. Results of the Theory of Mind trials are in Appendix Table \ref{tab:alignment_tom}. Column 11 shows the results of \texttt{DeepSeek-R1-70B}, the most advanced open-source reasoning model, as a robustness test.
        \end{tablenotes}
    \end{threeparttable}
\end{sidewaystable}

\begin{sidewaystable}[htbp]
    \caption{\textsc{LLM Agent's Alignment with Humans in Dictator Games: Compassion (Sense of Self)}} \label{tab:compassion_self}
    \begin{threeparttable}
        \begin{tabular}{rrccccccccccc}
            \hline\hline
              &              & (1)    & (2)    & (3)    & (4)    & (5)    & (6)    & (7)    & (8)    & (9)    & (10)   & Total \cmark \\
              &              & G27B   & G9B    & GPT4o  & L405B  & L70B   & L8B    & P14B   & P3.8B  & Q72B   & Q7B    & (by row)     \\
            \hline
            1 & Pos. Emotion & \cmark & \cmark & \cmark & \cmark & \cmark & n.s.   & \cmark & n.s.   & \cmark & \cmark & 8            \\
            2 & Social       & \cmark & \cmark & \cmark & \xmark & \cmark & n.s.   & n.s.   & n.s.   & n.s.   & n.s.   & 4            \\
            3 & Religious    & \xmark & \cmark & n.s.   & n.s.   & \cmark & n.s.   & n.s.   & n.s.   & n.s.   & \cmark & 3            \\
            4 & Affiliation  & \cmark & \cmark & \cmark & \cmark & \cmark & \cmark & \cmark & n.s.   & \cmark & \cmark & 9            \\
            5 & Certain      & \cmark & \xmark & \cmark & \xmark & \xmark & n.s.   & \xmark & n.s.   & \xmark & n.s.   & 2            \\
            6 & Family       & \xmark & \xmark & n.s.   & \xmark & \cmark & \xmark & \xmark & \cmark & n.s.   & \cmark & 3            \\
            7 & Drives       & \xmark & n.s.   & \xmark & \xmark & \xmark & \xmark & \xmark & n.s.   & \xmark & n.s.   & 0            \\
            8 & Affect       & \xmark & n.s.   & \xmark & \xmark & \cmark & \cmark & \xmark & n.s.   & \xmark & n.s.   & 2            \\
            \hline
              & Total \cmark & 4      & 4      & 4      & 2      & 6      & 2      & 2      & 1      & 2      & 4      & 31           \\
            \hline\hline
        \end{tabular}
        \begin{tablenotes}[para,flushleft]
            \footnotesize\textit{Note}: \cmark = Aligning with human studies; \xmark = Not aligning with human studies; n.s. = Not significant; pos. = Positive; neg. = Negative. ``--'' indicates the lack of consensus from human studies, showing directions of coefficients but not alignments for these variables. The expected directions of impact based on human studies are reviewed in Appendix \ref{sec:human_baseline_factors}. Results of the Theory of Mind trials are in Appendix Table \ref{tab:compassion_tom}.
        \end{tablenotes}
    \end{threeparttable}
\end{sidewaystable}

\begin{sidewaystable}[htbp]
    \caption{\textsc{LLM Agent's Alignment with Humans in Dictator Games: Empathy (Sense of Self)}} \label{tab:empathy_self}
    \begin{threeparttable}
        \begin{tabular}{rrccccccccccc}
            \hline\hline
               &                     & (1)    & (2)    & (3)    & (4)    & (5)    & (6)    & (7)    & (8)   & (9)    & (10)   & Total \cmark \\
               &                     & G27B   & G9B    & GPT4o  & L405B  & L70B   & L8B    & P14B   & P3.8B & Q72B   & Q7B    & (by row)     \\
            \hline
            1  & I                   & \cmark & \cmark & \cmark & \xmark & \xmark & n.s.   & \cmark & n.s.  & \xmark & \xmark & 4            \\
            2  & Focus Present       & \xmark & \xmark & \xmark & \xmark & \xmark & \xmark & \cmark & n.s.  & n.s.   & n.s.   & 1            \\
            3  & Personal Pronoun    & \xmark & n.s.   & \xmark & \xmark & \cmark & \cmark & \xmark & n.s.  & \xmark & n.s.   & 2            \\
            4  & Sadness             & n.s.   & n.s.   & \xmark & \cmark & n.s.   & \xmark & n.s.   & n.s.  & \xmark & n.s.   & 1            \\
            5  & Discrepancy         & n.s.   & \cmark & n.s.   & \cmark & \cmark & \cmark & \xmark & n.s.  & \xmark & \xmark & 4            \\
            6  & Verb                & \xmark & \xmark & \cmark & \xmark & \xmark & n.s.   & \xmark & n.s.  & \xmark & n.s.   & 1            \\
            7  & Adverb              & \cmark & \xmark & n.s.   & \cmark & n.s.   & \xmark & \xmark & n.s.  & \xmark & n.s.   & 2            \\
            8  & Cognitive Processes & \cmark & \cmark & \xmark & n.s.   & \xmark & \xmark & \xmark & n.s.  & n.s.   & n.s.   & 2            \\
            9  & Pronoun             & \cmark & \xmark & \cmark & \xmark & \xmark & \xmark & n.s.   & n.s.  & \xmark & n.s.   & 2            \\
            10 & Affect              & \xmark & n.s.   & \xmark & \xmark & \cmark & \cmark & \xmark & n.s.  & \xmark & n.s.   & 2            \\
            \hline
               & Total \cmark        & 4      & 3      & 3      & 3      & 3      & 3      & 2      & 0     & 0      & 0      & 21           \\
            \hline\hline
        \end{tabular}
        \begin{tablenotes}[para,flushleft]
            \footnotesize\textit{Note}: \cmark = Aligning with human studies; \xmark = Not aligning with human studies; n.s. = Not significant; pos. = Positive; neg. = Negative. ``--'' indicates the lack of consensus from human studies, showing directions of coefficients but not alignments for these variables. The expected directions of impact based on human studies are reviewed in Appendix \ref{sec:human_baseline_factors}. Results of the Theory of Mind trials are in Appendix Table \ref{tab:empathy_tom}.
        \end{tablenotes}
    \end{threeparttable}
\end{sidewaystable}

Tables \ref{tab:alignment_self}--\ref{tab:empathy_self} summarize the alignment of LLM agents with human behavior in dictator games under the Sense of Self perspective. The total number of \cmark\ marks in each column indicates the number of alignments with humans across all factors for a given model, reflecting the model's overall ability to be human-like. The total number of \cmark\ marks in each row indicates the number of alignments with humans for a given factor across all models, showing the overall consensus among different models on whether a factor \textit{should} aligns with human studies (i.e., ``industry consensus'').

In terms of being human-like, the \texttt{Llama3.1-405B} model demonstrates the highest total number of consistent results across all factors, aligning with human studies in 10 out of 14 factors, though no globally best model emerges. Surprisingly (or perhaps not, depending on how we frame the debiasing process in LLM development), the industry standard \texttt{GPT4o-2024-08-06} aligns with human studies in only two factors. For the alignment of psychological process, almost all models performed poorly. These results suggest that when LLM agents are instructed to adopt human personas, their behavior in the dictator game lacks clear patterns and exhibits significant inconsistencies. No consistent relationship emerges between their assigned personas and their decisions. Merely assigning a human-like identity to LLMs does not result in human-like behaviors.

Regarding which variable \textit{should} be an influencing factor, the models show the most consensus on \textit{Stranger Meet}---eight out of ten models suggest that if the dictator will meet the recipient after the game, they will behave more generously. For the alignment of psychological process, compassion-related processes represented by Positive Emotion and Affiliation (e.g., ``our,'' ``friends,'' ``family'') have the strongest consensus. Respectively, eight and nine out of ten models indicate that these processes should align with human studies.

Similarly, Appendix Tables \ref{tab:alignment_tom}--\ref{tab:empathy_tom} summarize the alignment of LLM agents with human behavior in dictator games under the Theory of Mind perspective, which closely resemble those of the Sense of Self trials. Two of the Llama3.1 models, \texttt{Llama3.1-405B} and \texttt{Llama3.1-70B}, exhibit the highest total number of consistent results across all factors, aligning with human studies in 10 out of 14 factors. The industry frontier model, \texttt{GPT4o-2024-08-06}, aligns with human studies in only 4 factors. In terms of psychological processes, the performance of LLM agents remains poor.\footnote{LIWC is probably not an appropriate method for estimating the reasoning process of these ToM trials. For example, these trials may use fewer first-person pronouns. Even when using these pronouns, their psychological meaning is different from that in the SoS trials.} These results suggest that when LLM agents are tasked with predicting human behavior based on their knowledge of humans, the results (Appendix \ref{sec:tom_results}) remain inconsistent and lack clear patterns. Despite being trained on extensive human-generated data, these AI agents cannot reason through human decision-making processes in dictator games.

These findings suggest that LLM agents' reasoning does not consistently exhibit textual markers of human decision-making in dictator games and that their alignment with human behavior varies substantially across model architectures and prompt formulations. The inconsistencies observed under both the Sense of Self and Theory of Mind perspectives highlight the limitations of LLMs to emulate human cognition and decision-making processes.

\clearpage
\section{Discussion}

Our study set out to examine whether LLMs can emulate or predict human behaviors in dictator games, a classic economic experiment designed to test the sense of fairness and altruism. By framing our research through the lenses of \textit{Sense of Self} and \textit{Theory of Mind} to test how persona assignments influence LLM behavior and whether LLMs can predict human decision-making, respectively, we aimed to understand the underlying mechanisms driving LLM decision-making and assess their alignment with human behaviors. The empirical results are summarized below:

\begin{enumerate}[topsep=0pt,itemsep=0pt,partopsep=0ex,parsep=0ex]
    \item \textit{Inconsistent Alignment with Human Behavior}: LLM agents did not consistently replicate human decision-making patterns in the dictator game. Assigning human-like personas or prompting them to predict human behavior did not result in outcomes that align with established human behaviors.
    \item \textit{Variability Across Models}: Significant variations exist both across different LLM families and within the same model family but different sizes. Larger models did not necessarily produce more human-like behaviors, and sometimes smaller models outperformed their larger counterparts in aligning with human.
    \item \textit{Lack of Continuous Decision Distribution}: Unlike humans, whose giving rates in dictator games typically follow a continuous distribution, LLM agents exhibited bimodal distributions, with choices clustered at extremes (e.g., giving nothing or half). This suggests a lack of nuanced decision-making that characterizes human prosocial behavior.
    \item \textit{Sensitivity to Experimental Framing}: While human decisions in dictator games are influenced by factors like social distance and framing (``Give'' vs. ``Take''), LLM agents showed inconsistent responses to these manipulations. Their behaviors did not consistently align with human expectations based on these contextual factors.
    \item \textit{Unpredictable Impact of Personas and Psychological Processes}: The assigned demographic and personality traits did not reliably predict the agents' decisions. Moreover, analyses of their textual explanations using LIWC did not reveal consistent psychological processes akin to human empathy or compassion.
\end{enumerate}

Two central themes emerge from these findings, highlighting some fundamental limitations and challenges of developing and applying LLMs in social contexts. The first theme pertains to \emph{what LLMs are actually learning}, and the second relates to \emph{how we should position LLMs within our society}.

\subsection{Inconsistency in LLM Behavior: Lack of Understanding and Theories}

The first theme highlights that current LLM agents do not consistently behave like humans in the specific context of dictator games. They appear to lack ``causal models of the world that support explanation and understanding'' and ``ground learning in intuitive theories of physics and psychology to support and enrich the knowledge that is learned'' \autocite[1]{LakeBuildingMachinesThat2017}. LLMs rely on recognizing language patterns rather than truly understanding social norms or engaging in human-like reasoning. Despite being trained on vast datasets of human-generated text, LLMs do not consistently replicate human decision-making in these social contexts. This inconsistency is further exacerbated by the models' sensitivity to factors such as architecture, size, and prompt formulations, which challenges the assumption that simply increasing model size or complexity inherently improves reasoning abilities or leads to more human-like behaviors.

While both LLMs and humans are epistemically opaque, there is a crucial difference. Human behaviors, though complex, can often be interpreted and predicted based on psychological theories and social norms. In contrast, LLMs lack such underlying theories; their internal processes remain a black box, and they do not follow human theories. This absence of interpretability and adherence to human reasoning processes limits our ability to understand and predict LLM behaviors in socially complex scenarios.

\subsection{Determinism vs. Human-Like Uncertainty: A Fundamental Dilemma} \label{sec:determinism}

The second theme centers on the dichotomy between deterministic outputs and human-like uncertainty in LLM behavior. The bimodal distribution of giving rates among LLM agents suggests a form of deterministic decision-making that lacks the subtlety and variability characteristic of human choices. While deterministic behavior might result in more predictable outputs suitable for certain applications, it fails to capture the richness of human behavior, which often involves nuanced deliberation over various social and personal factors.

The absence of a continuous decision space indicates that LLMs may be defaulting to prevalent patterns in their training data or adhering to the most statistically probable responses. This tendency suggests that they are not genuinely understanding or processing the ethical dimensions of the choices presented to them but are instead relying on learned language patterns. This brings us to a fundamental question: \emph{Should LLMs be designed to mimic human-like uncertainty, embracing the complexities and unpredictabilities of human decision-making, or should they aim for determinism to ensure consistency and predictability?}

This dilemma has significant implications for the development and deployment of LLMs. On one hand, embracing human-like uncertainty could enhance the authenticity of interactions with AI agents, making them more relatable and better suited for applications requiring empathy and nuanced social understanding. On the other hand, deterministic behavior ensures reliability and predictability, which are crucial for tasks where consistency is key.

\subsection{LLMs being Human-Like, but ``Which Human?''} \label{rl:which_human}

A critical assumption in this study is that a ``ground truth'' for typical human behavior can be established by summarizing the consensus from existing scholarship on dictator games. While this approach provides a necessary baseline for comparison, it opens a crucial line of inquiry succinctly captured by \textcite{AtariWhichHumans2023}: when we evaluate an LLM against ``human'' performance, \emph{which humans} are we talking about? The notion of a ``typical'' human is highly contested, as much of the behavioral science literature is dominated by research on participants from \emph{Western, Educated, Industrialized, Rich, and Democratic (WEIRD)} societies \autocite{HenrichWeirdestPeopleWorld2010}. This narrow sampling raises questions about the generalizability of findings concerning core human behaviors like altruism and fairness. \citeauthor{AtariWhichHumans2023} demonstrated that LLMs' psychological profiles most closely resemble those of people from WEIRD cultures, and this resemblance decreases as one moves away from that demographic.

The ``Which Human'' question highlights a fundamental challenge of representativeness, which can be examined from three perspectives: the training data, the academic knowledge base, and the resulting language model behavior.

First, the \emph{training data} used for LLMs is heavily skewed. These models are trained on vast quantities of text from the internet, a space where content from North America and Europe is disproportionately represented. Consequently, the social norms, biases, and behavioral patterns the LLM learns are not representative of global human diversity but rather reflect a digitally dominant, often WEIRD, slice of humanity. The model's foundational understanding of ``prosocial behavior'' is therefore culturally biased from the outset.

Second, the \emph{academic knowledge} that forms our human baseline suffers from the same bias. The empirical studies and literature reviews summarized in this paper to define expected human behavior (Appendix \ref{sec:human_baseline_factors}) are themselves largely products of research conducted within WEIRD populations. Therefore, our experiment compares an LLM trained on WEIRD-centric data to a behavioral benchmark derived from WEIRD-centric science. This framework risks reinforcing a culturally specific model of behavior as a universal standard.

Finally, these issues converge in the \emph{language model's behavior}. The inconsistencies and deviations from the human baseline observed in our results may not simply be technical failures of the models. Instead, they could reflect a complex conflict between the LLM's core WEIRD-centric training, the specific (U.S.-based) personas it is asked to adopt, and the debiasing efforts intended to make it a more neutral agent. Such debiasing may strip the model of the ability to replicate any specific human demographic's patterns faithfully, resulting in the sanitized and inconsistent outputs we observed, particularly with a highly-tuned model like GPT-4o. Future work on prosocial AI must therefore move beyond simple alignment with a monolithic ``human'' standard and instead grapple with the challenge of building AI that can understand and navigate the rich diversity of human cultural and social norms.

\subsection{Practical Implications for Developing and Deploying LLMs}

\textit{Behavioral Approach to Evaluating Internal Processes of LLMs.} Our study underscores the challenges in aligning LLM behaviors with human values and social norms, highlighting the need for more sophisticated evaluation methods. Traditional approaches that focus on adjusting outputs based on human feedback are insufficient for tasks requiring social cognition and reasoning. As discussed earlier, adopting a behavioral approach---such as evaluating LLMs through experiments---allows us to systematically assess their decision-making processes in realistic social contexts. This method provides insights into how LLMs make decisions and whether their internal mechanisms align with human cognitive processes.

\textit{Assistants for Tasks but Not Participants in Social Research.} The use of LLMs in social science research is promising but also presents limitations. Our findings suggest that LLMs may not reliably replicate the nuanced processes of human decision-making in social experiments like the dictator game---they are not computational humans. Over-relying on them for modeling human behavior in complex social contexts could lead to misleading conclusions. This is particularly relevant for the nonprofit and philanthropic sectors, where AI might be used to model donor behavior or predict responses to fundraising campaigns. Inaccurate simulations could lead to flawed strategies and misallocation of resources. Therefore, researchers and practitioners should limit the roles of LLMs to specific tasks like text classification or topic modeling and approach the use of LLMs in modeling human behavior with caution. We must recognize that LLMs are tools to assist in research, not substitutes for human participants, at least for the time being.

~\\

\noindent As society increasingly relies on AI for critical decision-making tasks, integrating prosocial AI into NPS becomes both timely and imperative. This study highlights the limitations and opportunities associated with LLMs' prosocial behaviors, underscoring the importance of interdisciplinary collaboration between computer science, traditional social sciences, and philanthropic studies. NPS are uniquely positioned, through their extensive understanding of human behaviors, ethics, and societal norms, to guide the development and application of prosocial AI technologies, ensuring these systems align with the core values of human society and the practical needs of nonprofit and philanthropic sectors.





\clearpage
\begingroup
\singlespacing
\sloppy
\printbibliography[title={References}]
\endgroup

\clearpage
\begingroup
\section*{\textsc{Online Appendix}}
\begin{appendix}
  \begin{refsection}

\renewcommand\thetable{\Alph{section}\arabic{table}}
\renewcommand\thefigure{\Alph{section}\arabic{figure}}
\setcounter{footnote}{0}
\setcounter{page}{1}

\begin{center}
    \singlespacing\textsc{\mytitle}\\~\\
\end{center}

\vspace{-2cm}

\begingroup
\onehalfspacing
\normalsize
\etocdepthtag.toc{mtappendix}
\etocsettagdepth{mtchapter}{none}
\etocsettagdepth{mtappendix}{subsubsection}
\setcounter{tocdepth}{3}
\tableofcontents
\endgroup

\clearpage

\section{Literature Reivew and Knowledge Base}

\subsection{Evaluating LLMs as Tools for Specific Tasks} \label{sec:llm_as_tools}

\subsubsection{Benchmarks in Computer Science} \label{sec:cs_benchmarks}

In computer science and computational linguistics, benchmarks have been instrumental in evaluating the performance of language models. Early benchmarks focused on specific, well-defined tasks such as part-of-speech tagging, named entity recognition, and syntactic parsing. As language models evolved, so did the benchmarks, leading to more comprehensive evaluations that test a model's understanding and reasoning capabilities.

A significant milestone was the introduction of the General Language Understanding Evaluation (GLUE) benchmark \autocite{WangGLUEMultiTaskBenchmark2019}. GLUE was designed to promote the development of generalizable natural language understanding systems. The benchmark was structured so that achieving good performance would require a model to share substantial knowledge across all tasks while still maintaining some task-specific components. GLUE aggregates nine English sentence understanding tasks, such as sentiment analysis, textual entailment, and question-answering. As models began to surpass non-expert human performance on GLUE, the SuperGLUE benchmark was proposed \autocite{WangSuperGLUEStickierBenchmark2020}, offering more challenging tasks that require advanced reasoning and world knowledge.

Large-scale language models like GPT-3 significantly pushed the boundaries of what benchmarks needed to assess \autocite{BrownLanguageModelsAre2020}. These LLMs demonstrated impressive zero-shot and few-shot learning capabilities, handling a variety of tasks without explicit training on them. Consequently, more recent benchmarks have aimed to evaluate models across an even wider range of tasks. The Massive Multitask Language Understanding (MMLU) benchmark assesses models on 57 tasks spanning mathematics, humanities, sciences, and more, testing their breadth of knowledge and reasoning skills \autocite{HendrycksMeasuringMassiveMultitask2021}. Similarly, the BIG-bench project encompasses an extensive collection of 204 tasks contributed by 450 authors across 132 institutions \autocite{SrivastavaImitationGameQuantifying2023}. The tasks are diverse, covering areas such as linguistics, childhood development, mathematics, common-sense reasoning, biology, physics, social bias, software development, and beyond.

Despite this impressive breadth, most benchmarks still share fundamental limitations. First, the questions within these benchmarks are not open-ended, which hinders the ability to capture the flexible and interactive use of language found in real-world settings. Second, for many complex tasks, establishing a definitive ground truth is challenging or sometimes unattainable. As a result, current benchmarks fail to adequately address the needs of state-of-the-art (SOTA) LLMs, particularly in evaluating user preferences \autocite[1]{ChiangChatbotArenaOpen2024}. Finally, the data from benchmark tests can become part of the training datasets for newer models, rendering these benchmarks obsolete. Such test set contamination is particularly problematic for LLMs, which are trained on vast amounts of online data \autocite{WhiteLiveBenchChallengingContaminationFree2024}. There is an urgent need for open, live evaluation platforms based on human preferences that can more accurately mirror real-world usage. Platforms like Chatbot Arena, Arena-Hard, and LiveBench address this by enabling live evaluations where users can interact with different language models in real-time conversations and vote for the best models according to their own preferences, allowing assessments in more naturalistic and uncontaminated settings \autocites{ChiangChatbotArenaOpen2024}{WhiteLiveBenchChallengingContaminationFree2024}{LiCrowdsourcedDataHighQuality2024}.

While numerous other benchmarks have been developed for various purposes---far beyond the scope of this paper to detail---they remain largely task-specific and context-free. Moreover, these benchmarks mainly focus on comparing final outputs without providing insights into the internal decision-making processes of LLMs, how these processes are influenced by various factors, or how they compare to human cognition. As \textcite{BenderClimbingNLUMeaning2020} argue, evaluations should test models on their understanding of the world and language use in context rather than just on form-based tasks.

\subsubsection{``Text as Data'' in Social Sciences} \label{text_as_data}

In social sciences, analyzing ``text as data'' with advanced computational methods to study human behavior and social phenomena has become a well-established approach \autocites{GrimmerTextDataPromise2013}{GrimmerTextDataNew2022}. Researchers have employed text analysis methods on large volumes of textual data from various sources to study a variety of topics, such as political behavior \autocite{RobertsIntroductionVirtualIssue2016}, organizational research \autocites{KobayashiTextMiningOrganizational2018}{HickmanTextPreprocessingText2022}, and psychological processes \autocites{TausczikPsychologicalMeaningWords2010}{YadenCharacterizingEmpathyCompassion2024}. In these social science studies, text analysis methods and algorithms are commonly used as tools to help researchers identify patterns or code empirical data into theoretical categories.

For example, researchers quantify important social constructs---such as social stereotypes \autocite{JonesStereotypicalGenderAssociations2020}, culture \autocite{KozlowskiGeometryCultureAnalyzing2019}, and the formation of scientific consensus \autocite{MaConsensusFormationNonprofit2024}---using text data and word embeddings \autocite{RodriguezWordEmbeddingsWhat2022}. They also automate the coding of text data into theoretical categories, such as political sentiments and stances \autocites{YoungAffectiveNewsAutomated2012}{BestvaterSentimentNotStance2023}, and the priorities and reputations of administrative bureaucracies \autocites{HollibaughUseTextData2019}{AnastasopoulosMachineLearningPublic2019}, using machine learning algorithms. Additionally, unsupervised topic modeling can be employed to advance social and management theories \autocites{BaumerComparingGroundedTheory2017}{HanniganTopicModelingManagement2019}.

With the development of LLMs, the potential for processing and analyzing text data in social science has expanded significantly. Due to their zero-shot and few-shot learning capabilities---which allow them to excel in specific tasks without extensive manually compiled training data or with only a very small training dataset---LLMs can annotate text data in social science research without the need for extensive manual coding or labeling \autocite{ZiemsCanLargeLanguage2024}. Beyond conventional coding tasks, scholars also found that LLMs have an impressive ability to generate novel research ideas and testable hypotheses based on existing scholarship \autocites{BankerMachineassistedSocialPsychology2024}{ZhouHypothesisGenerationLarge2024}, further raising emergent questions about how LLMs can improve or reshape social science research \autocites{BailCanGenerativeAI2024}{KozlowskiSimulatingSubjectsPromise2024}{Chang12BestPractices2024}.

From the initial application of simple algorithms to the current use of advanced LLMs, scientists have primarily employed these AI tools for specific tasks with clear objectives, such as classifying text data into predefined categories and extracting topics. These tasks are well-defined and come with clear benchmarks for evaluation, with human validation typically recommended as the standard to assess the performance of these algorithms \autocite{GrimmerTextDataPromise2013}. Even though humans make mistakes, they are still considered the ``gold standard'' \autocite{SongValidationsWeTrust2020}.

\subsection{Human Baseline: Influencing Factors in Dictator Games} \label{sec:human_baseline_factors}

Understanding human generosity requires exploring a complex interplay of factors, including demographics, personality traits, and social context. These elements are often studied through experimental methods like the dictator game, ultimatum game, and public goods game, with the dictator game being particularly popular among researchers \autocites{EngelDictatorGamesMeta2011}{ListInterpretationGivingDictator2007}. In a typical dictator game, one participant (the dictator) is given a certain amount of money or resources and must decide how much, if any, to share with another participant (the recipient), who has no power to influence the decision. This experimental setup provides valuable insights into the factors that drive altruistic behavior in a controlled environment.

Research has identified several factors that consistently influence generosity in dictator games (Table \ref{tab:generosity_factors}). Demographic factors such as age, gender, economic status, and education significantly impact individuals' decisions to give. Personality traits, particularly Agreeableness and Openness, are crucial in shaping generosity. The framing of experiments, such as the level of social distance or the specific nature of the giving scenario, also influences prosocial behavior. Additionally, psychological mechanisms like compassion and empathy motivate individuals to act generously, each involving distinct emotional and cognitive processes. This section reviews these factors, primarily studied through dictator games, to provide an overview of what drives generosity in human behavior.

\begin{sidewaystable}[htbp]
    \caption{\textsc{Factors Influencing Human Generosity in dictator games}} \label{tab:generosity_factors}
    \small
    \begin{threeparttable}
        \begin{tabular}{r>{\raggedright\arraybackslash}p{9cm}>{\raggedright\arraybackslash}p{9cm}}
            \hline\hline
            \multicolumn{1}{l}{}                               & Empirical Research                                                                                                                                                                                                      & General Consensus                                                                                                                                                                                                            \\
            \hline

                                                               &                                                                                                                                                                                                                         &                                                                                                                                                                                                                              \\
            \multicolumn{1}{l}{\textit{Demographics}}          &                                                                                                                                                                                                                         &                                                                                                                                                                                                                              \\
            Age                                                & \textcites{BekkersMeasuringAltruisticBehavior2007}{EngelDictatorGamesMeta2011}{MatsumotoProsocialBehaviorIncreases2016}                                                                                                 & Positive effect on giving.                                                                                                                                                                                                   \\
            Gender                                             & \textcites{EngelDictatorGamesMeta2011}{EaglyHisHersProsocial2009}{SaadEffectsRecipientsGender2001}{BekkersMeasuringAltruisticBehavior2007}{Donate-BuendiaGenderOtherModerators2022}                                     & Females give more.                                                                                                                                                                                                           \\
            Economic status                                    & \textcites{MacchiaLinkIncomeIncome2021}{CappelenGiveTakeDictator2013}{CochardSocialPreferencesDifferent2021}{ChenFamilyIncomeAffects2013}{RaoDisadvantageProsocialBehavior2011}{BekkersMeasuringAltruisticBehavior2007} & Lack of consensus: At country level, participants from high-income countries give less than those from low-income countries; at individual level, findings about the relation between income and generosity are conflicting. \\
            Education                                          & \textcite{BekkersMeasuringAltruisticBehavior2007}                                                                                                                                                                       & Positive effect on giving.                                                                                                                                                                                                   \\
            Marriage                                           & \textcites{BekkersWhoGivesLiterature2011}{TwengeSocialExclusionDecreases2007}                                                                                                                                           & Positive effect on giving.                                                                                                                                                                                                   \\

                                                               &                                                                                                                                                                                                                         &                                                                                                                                                                                                                              \\
            \multicolumn{1}{l}{\textit{Personality}}           &                                                                                                                                                                                                                         &                                                                                                                                                                                                                              \\
            Big Five                                           & \textcites{HabashiSearchingProsocialPersonality2016}{KlinePersonalityProsocialBehavior2019}{FurnhamBigFiveBig1996}                                                                                                      & Agreeableness and Openness have positive effect on giving.                                                                                                                                                                   \\

                                                               &                                                                                                                                                                                                                         &                                                                                                                                                                                                                              \\
            \multicolumn{1}{l}{\textit{Experiment Framing}}    &                                                                                                                                                                                                                         &                                                                                                                                                                                                                              \\
            Social distance                                    & \textcites{EngelDictatorGamesMeta2011}{Goeree1LawGiving2010}{BechlerProportionOfferedDictator2015}                                                                                                                      & Individuals give more to closer friends.                                                                                                                                                                                     \\
            Give vs. Take                                      & \textcites{ListInterpretationGivingDictator2007}{CappelenGiveTakeDictator2013}{BardsleyDictatorGameGiving2008}{KorenokTakingGivingImpure2014}                                                                           & Take option significantly reduces giving.                                                                                                                                                                                    \\
            Stake                                              & \textcite{EngelDictatorGamesMeta2011}                                                                                                                                                                                   & Higher stakes reduce both the absolute amount and the percentage of giving.                                                                                                                                                  \\

                                                               &                                                                                                                                                                                                                         &                                                                                                                                                                                                                              \\
            \multicolumn{1}{l}{\textit{Psychological Process}} &                                                                                                                                                                                                                         &                                                                                                                                                                                                                              \\
            Compassion                                         & \textcites{TausczikPsychologicalMeaningWords2010}{YadenCharacterizingEmpathyCompassion2024}                                                                                                                             & Focuses on social relations, others, and positive emotions.                                                                                                                                                                  \\
            Empathy                                            & \textcites{TausczikPsychologicalMeaningWords2010}{YadenCharacterizingEmpathyCompassion2024}                                                                                                                             & Focuses on self, negative emotions, and present.                                                                                                                                                                             \\
                                                               &                                                                                                                                                                                                                         &                                                                                                                                                                                                                              \\
            \hline\hline
        \end{tabular}
        \begin{tablenotes}[flushleft]
            \footnotesize
            \item\noindent\footnotesize{\emph{Note}: Both review articles and empirical studies are included. Review articles are prioritized when consensus is strong.}
        \end{tablenotes}
    \end{threeparttable}
\end{sidewaystable}

\subsubsection{Demographics}

\textbf{Age.} Research indicates that generosity tends to increase with age. \textcite[139]{BekkersMeasuringAltruisticBehavior2007} found that generosity positively correlates with several factors, including age, education, income, trust, and a prosocial value orientation, in dictator games. \textcite[599]{EngelDictatorGamesMeta2011}'s meta-analysis of empirical studies also supports this, demonstrating a strong, statistically significant effect of age on generosity in dictator games, where older individuals exhibit higher levels of generosity compared to younger ones. The positive relationship between age and prosocial behavior is also widely observed beyond dictator games in many popular economic games \autocite{MatsumotoProsocialBehaviorIncreases2016}. This increase in generosity with age may be attributed to greater life experience, increased empathy, and a more established sense of social responsibility among older individuals.

\textbf{Gender.} Research consistently shows that gender differences influence generosity in dictator games, with females typically giving more than males as dictators \autocites[597]{EngelDictatorGamesMeta2011}{EaglyHisHersProsocial2009} and also receiving more as recipients \autocite{SaadEffectsRecipientsGender2001}. Women tend to engage in more prosocial behaviors that are communal and relational, whereas men are more inclined toward agentic, strength-intensive behaviors; and the origins of these differences may lie in traditional divisions of labor and biosocial interactions related to gender roles \autocite{EaglyHisHersProsocial2009}. A comprehensive meta-analysis of existing studies found that these gender differences persist across various experimental conditions and locations, with women generally being more generous than men. However, women are less generous than men when interacting with close friends or family members, indicating that the context and social distance can modulate these gender effects \autocite{Donate-BuendiaGenderOtherModerators2022}. Overall, while women exhibit greater generosity in many scenarios, the influence of social norms and situational factors remains significant.

\textbf{Economic status.} The relationship between economic status and generosity is mixed and varies depending on the level of analysis and context \autocite[375--376]{MacchiaLinkIncomeIncome2021}. At the country level, lower-income countries tend to exhibit higher levels of generosity, with studies indicating that participants from these countries are more likely to give away a greater proportion of resources compared to those from higher-income countries \autocite[595]{CappelenNeedsEntitlementsInternational2013}. This may be due to a stronger adherence to fairness norms in less economically developed nations \autocite[1]{CochardSocialPreferencesDifferent2021}. At the individual level, some studies, such as those by \textcite{BekkersMeasuringAltruisticBehavior2007} and \textcite{MacchiaLinkIncomeIncome2021}, found that higher-income individuals are more likely to donate money and volunteer their time. However, other studies, like \textcite{ChenFamilyIncomeAffects2013}, found that children from lower-income families displayed more altruistic behavior, possibly due to local socialization practices. Additionally, catastrophic events, such as the 2008 earthquake in China, can temporarily increase prosocial behavior among those directly affected, reflecting a contextual impact on generosity \autocite{RaoDisadvantageProsocialBehavior2011}. Overall, while higher income may correlate with increased giving, context and local social norms play crucial roles in shaping prosocial behaviors across different economic strata.

\textbf{Education}, though not commonly examined in empirical studies using dictator games \autocite{EngelDictatorGamesMeta2011}, has consistently shown a positive influence on generosity in broader studies of prosocial behavior. \textcite{BekkersMeasuringAltruisticBehavior2007} found that more educated individuals tend to give more in dictator games, likely because those with higher education levels have a greater awareness of need and a stronger alignment with prosocial values \autocite[344--349]{BekkersWhoGivesLiterature2011}.

\textbf{Marriage} has a positive influence on generosity. According to \textcite{BekkersWhoGivesLiterature2011}, married individuals tend to be more generous, possibly due to the increased social networks and responsibilities associated with marriage. Additionally, \textcite{TwengeSocialExclusionDecreases2007} found that social connectedness, which is often stronger in married individuals, can lead to greater prosocial behaviors, such as charitable giving, volunteering, and cooperation in social settings.

\subsubsection{Personality}

Personality traits also play a notable role in influencing generosity. Research has shown that among the Big Five personality traits, Agreeableness is most closely associated with positive emotional reactions to individuals in need and subsequent decisions to help \autocite[1177]{HabashiSearchingProsocialPersonality2016}. Additionally, both Agreeableness and Openness are significantly and positively related to prosocial behavior, while the other three traits (Conscientiousness, Extraversion, Neuroticism) show no such relationship \autocite[125]{KlinePersonalityProsocialBehavior2019}. Given the strong correlation between the MBTI and Big Five personality traits \autocites{FurnhamBigFiveBig1996}[127]{KlinePersonalityProsocialBehavior2019}, individuals with MBTI types characterized by Extraversion (E), Intuition (N), Feeling (F), and Perceiving (P) are likely to be more generous, aligning with the traits of Agreeableness and Openness.

\subsubsection{Experiment Framing}

\textbf{Social distance} in dictator games refers to the perceived closeness or relationship between the dictator and the recipient. Manipulations of social distance can include varying the anonymity of participants or providing personal information about the recipient. Existing studies show that the proportion of giving decreases as social distance increases---in other words, people tend to give more to close friends than to distant strangers. See empirical results from \textcite[192, Figure 2]{Goeree1LawGiving2010}, \textcite[152, Figure 1]{BechlerProportionOfferedDictator2015}, and a meta-analysis of \textcite[597, Figure 7]{EngelDictatorGamesMeta2011}. In all these studies, a smaller social distance value between the dictator and recipient indicates a closer relationship.

\textbf{Give vs. Take.} The framing of choices in dictator games can influence the dictator's generosity, particularly in how the choice is presented as either giving or taking. In the ``Give'' framing, dictators decide how much of their endowment to give away, while in the ``Take'' framing, they have the opportunity to take away from the recipient's initial endowment. Studies have found that the inclusion of a ``Take'' option significantly reduces the amount transferred to recipients; in addition, different framings regarding whether the recipients are ``entitled'' (e.g., earned versus unearned income) to their initial endowment can also significantly impact the amount transferred \autocites{ListInterpretationGivingDictator2007}{CappelenGiveTakeDictator2013}{BardsleyDictatorGameGiving2008}{KorenokTakingGivingImpure2014}.

\textbf{Stake.} The amount of money at stake in the dictator game can also impact generosity. Higher stakes are associated with a reduced willingness to give; when there is more to gain, dictators tend to keep more for themselves, both in absolute and relative terms \autocite[592]{EngelDictatorGamesMeta2011}.

\subsubsection{Psychological Process}

The psychological processes of compassion and empathy are fundamental in shaping prosocial behaviors. Empathy involves feeling what we believe others are feeling, which allows us to emotionally connect with their experiences. Compassion, on the other hand, involves caring for and about others without necessarily sharing their feelings, focusing more on a desire to help and alleviate suffering. Using the Linguistic Inquiry and Word Count \autocite{TausczikPsychologicalMeaningWords2010}, \textcite{YadenCharacterizingEmpathyCompassion2024} analyzed over two million Facebook posts from 2.7 thousand individuals and found that those high in empathy often use self-focused language and discuss negative emotions and social isolation. In contrast, individuals high in compassion use other-focused language, expressing positive feelings and social connections. The study also found that high empathy without compassion is linked to negative health outcomes, while high compassion without empathy is associated with positive health outcomes, healthy lifestyle choices, and charitable giving. These findings suggest that compassion, rather than empathy, might be a more effective driver of prosocial behavior and moral motivation.

\clearpage
\section{Selection of LLMs} \label{sec:llm_selection}

Our selection criteria for the LLMs included: 1. Open-source foundation models \autocite{BommasaniOpportunitiesRisksFoundation2022}, chosen for their transparency, reproducibility, and widespread use \autocites{SpirlingWhyOpensourceGenerative2023}{BailCanGenerativeAI2024}; 2. Models demonstrating SOTA performance \autocite{FourrierOpenLLMLeaderboard2024}, ensuring we capture the highest achievable and quality results; 3. Models released by multinational and leading technology companies, as these models are likely to be embedded in widely used products (e.g., Microsoft Word and Gmail) and can potentially reach millions, if not billions, of users. Based on these criteria, we selected the following model families for our experiments, testing both the smallest and largest size models within each family:

\begin{enumerate}[topsep=0pt,itemsep=0pt,partopsep=0ex,parsep=0ex]

    \item \textit{Llama3.1}\footnote{\url{https://www.llama.com/}}: Developed by Meta (the parent company of Facebook) and released on July 23, 2024, this model series consistently achieves SOTA results in many areas, such as reasoning, coding, and multilingual abilities, serving as a benchmark for other open-source foundation models. This study uses Llama 3.1 models in 8B, 70B, and 405B (B = Billion).

    \item \textit{Qwen2.5}\footnote{\url{https://github.com/QwenLM/Qwen2.5}}: Released by the Qwen team from Alibaba Cloud on September 19, 2024. Alibaba Cloud is a subsidiary of Alibaba Group and one of the largest cloud computing providers globally. This model family is multilingual, specializing in English and Chinese and supporting 29 languages. It achieves results comparable to the Llama family models on various tasks \autocite{FourrierOpenLLMLeaderboard2024}. The two model sizes used in this study are 7B and 72B.

    \item \textit{Gemma2}\footnote{\url{https://ai.google.dev/gemma}}: Engineered by Google and released on Jun 27, 2024, the Gemma2 series focuses on efficiency and performance \autocite{GemmaTeamGemma2Improving2024}. We tested Gemma 2 models in 9B and 27B.

    \item \textit{Phi3}\footnote{\url{https://microsoft.com/phi3}}: Microsoft released this model family on April 23, 2024. Phi3 models are tailored for small devices, such as smartphones, combining compactness with powerful computational abilities \autocite{AbdinPhi3TechnicalReport2024}. We tested Phi3 models in 3.8B and 14B.

    \item \textit{GPT4o}\footnote{\url{https://openai.com/index/hello-gpt-4o/}}: One of the most advanced models developed by OpenAI. While the model is proprietary and its exact size is undisclosed, it is widely recognized as the current SOTA for all LLMs and is commonly held as the highest industry standard. We tested the GPT4o model released in August 2024 (``2024-08-06'') in this study.

\end{enumerate}

Depending on their target applications, these models vary not only in size but also in architecture. In general, the larger models are more capable but are computationally demanding, while the smaller models are more lightweight and suitable for devices with limited resources.

\clearpage
\section{Prompts for LLM Agents and Textual Output of Reasoning} \label{sec:prompts}
\setcounter{table}{0}
\setcounter{figure}{0}

\begin{lstlisting}[language=json,firstnumber=1,caption={\textsc{Agent Setting Instruction: Sense of Self}}, label={prom:agent_setting_instruction_self}]
You are an individual living in the United States with this profile:
{dictator_profile}.
Always think step by step.
Only return your response in json format with "amount_transfer", "reason_transfer", "final_payment_you", "final_payment_other" keys.
Dollar amounts only in positive or negative numbers.
Give final amounts, don't show the calculation.
Don't add + to positive numbers.
Don't add dollar sign or dollar unit to the amounts.
\end{lstlisting}
\vspace{-1em}
\begin{minipage}{\linewidth}
    \footnotesize{\emph{Note}: Variables randomly sampled each trial are indicated with curly brackets (``\{\}'').}
\end{minipage}

\begin{lstlisting}[language=json,firstnumber=1,caption={\textsc{Agent Setting Instruction: Theory of Mind}}, label={prom:agent_setting_instruction_tom}]
You predict the behavior of a decision-maker in a dictator game according to your knowledge about human behavior.
The decision-maker is a human individual living in the United States with this profile: {dictator_profile}.
Always think step by step.
Only return your response in JSON format with "amount_transfer", "reason_transfer", "final_payment_dmaker", "final_payment_recipient" keys.
Dollar amounts only in positive or negative numbers.
Give final amounts, don't show the calculation.
Don't add + to positive numbers.
Don't add dollar sign or dollar unit to the amounts.
\end{lstlisting}
\vspace{-1em}
\begin{minipage}{\linewidth}
    \footnotesize{\emph{Note}: Variables randomly sampled each trial are indicated with curly brackets (``\{\}'').}
\end{minipage}

\clearpage
\begin{lstlisting}[language=json,firstnumber=1,caption={\textsc{Game Instruction: Sense of Self, ``Give'' Framing}}, label={prom:game_instruction_self_give}]
You are now paired with another participant.
{social_distance_dict[social_distance]}
Both of you have been allocated {amount_given} USD in this part of the experiment.
In addition, you have been provisionally allocated an additional {amount_given} USD.
The other participant has NOT been allocated these additional {amount_given} USD.
Your decision is a simple one: Decide what portion, if any, of these {amount_given} USD to transfer to the other person.
You can choose any amount from 0 USD to {amount_given} USD to transfer.
Your payment is your initial {amount_given} USD allocation plus the amount that is allocated to you given your decision.
The other participant's payment is his or her initial {amount_given} USD plus the amount that follows from your decision.
The other person will not make any decision, but he or she has the opportunity to read the instructions we have given to you.
How much would you like to transfer to the other participant ("amount_transfer"), and why ("reason_transfer")?
How much is your final payment ("final_payment_you"), and how much is the other person's final payment ("final_payment_other")?
\end{lstlisting}
\vspace{-1em}
\begin{minipage}{\linewidth}
    \footnotesize{\emph{Note}: Variables randomly sampled each trial are indicated with curly brackets (``\{\}'').}
\end{minipage}

\begin{lstlisting}[language=json,firstnumber=1,caption={\textsc{Game Instruction: Sense of Self, ``Take'' Framing}}, label={prom:game_instruction_self_take}]
You are now paired with another participant.
{social_distance_dict[social_distance]}
Both of you have been allocated {amount_given} USD in this part of the experiment.
In addition, you have been provisionally allocated an additional {amount_given} USD.
The other participant has not been allocated these additional {amount_given} USD.
Your decision is a simple one: Decide what portion, if any, of these {amount_given} USD to transfer to the other person.
You can also transfer a negative amount. This means that you can take up to {amount_given} USD from the other participant.
You can choose any amount from -{amount_given} USD to {amount_given} USD to transfer.
Your payment is your initial {amount_given} USD allocation plus the amount that is allocated to you given your decision.
The other participant's payment is his or her initial {amount_given} USD plus the amount that follows from your decision.
The other person will not make any decision, but he or she has the opportunity to read the instructions we have given to you.
How much would you like to transfer to the other participant ("amount_transfer"), and why ("reason_transfer")?
How much is your final payment ("final_payment_you"), and how much is the other person's final payment ("final_payment_other")?
\end{lstlisting}
\vspace{-1em}
\begin{minipage}{\linewidth}
    \footnotesize{\emph{Note}: Variables randomly sampled each trial are indicated with curly brackets (``\{\}'').}
\end{minipage}

\clearpage
\begin{lstlisting}[language=json,firstnumber=1,caption={\textsc{Game Instruction: Theory of Mind, ``Give'' Framing}}, label={prom:game_instruction_tom_give}]
A decision-maker is paired with a recipient.
{social_distance_dict[social_distance]}
Both of them have been allocated {amount_given} USD in this part of the experiment.
In addition, the decision-maker have been provisionally allocated an additional {amount_given} USD.
The recipient has NOT been allocated these additional {amount_given} USD.
The decision-maker needs to make a decision: What portion, if any, of these additional {amount_given} USD to transfer to the recipient.
The decision-maker can choose any amount from 0 USD to {amount_given} USD to transfer.
The payment to the decision-maker is the initial {amount_given} USD allocation plus the amount that follows from the decision.
The recipient's payment is his or her initial {amount_given} USD plus the amount that follows from the decision.
The recipient will not make any decision, but both the decision-maker and recipient read the instructions.
How much should the decision-maker transfer to the recipient ("amount_transfer"), and why ("reason_transfer")?
How much is the decision-maker's final payment ("final_payment_dmaker"), and how much is recipient's final payment ("final_payment_recipient")?
\end{lstlisting}
\vspace{-1em}
\begin{minipage}{\linewidth}
    \footnotesize{\emph{Note}: Variables randomly sampled each trial are indicated with curly brackets (``\{\}'').}
\end{minipage}

\begin{lstlisting}[language=json,firstnumber=1,caption={\textsc{Game Instruction: Theory of Mind, ``Take'' Framing}}, label={prom:game_instruction_tom_take}]
A decision-maker is paired with a recipient.
{social_distance_dict[social_distance]}
Both them have been allocated {amount_given} USD in this part of the experiment.
In addition, the decision-maker have been provisionally allocated an additional {amount_given} USD.
The recipient has NOT been allocated these additional {amount_given} USD.
The decision-maker needs to make a decision: What portion, if any, of these additional {amount_given} USD to transfer to the recipient.
The decision-maker can also transfer a negative amount. This means that the decision-maker can take up to {amount_given} USD from the recipient.
The decision-maker can choose any amount from -{amount_given} USD to {amount_given} USD to transfer.
The payment to the decision-maker is the initial {amount_given} USD allocation plus the amount that follows from the decision.
The recipient's payment is his or her initial {amount_given} USD plus the amount that follows from the decision.
The recipient will not make any decision, but both the decision-maker and recipient read the instructions.
How much should the decision-maker transfer to the recipient ("amount_transfer"), and why ("reason_transfer")?
How much is the decision-maker's final payment ("final_payment_dmaker"), and how much is recipient's final payment ("final_payment_recipient")?
\end{lstlisting}
\vspace{-1em}
\begin{minipage}{\linewidth}
    \footnotesize{\emph{Note}: Variables randomly sampled each trial are indicated with curly brackets (``\{\}'').}
\end{minipage}

\clearpage
\begin{lstlisting}[language=json,firstnumber=1,caption={\textsc{Example Textual Output of Reasoning (1)}}, label={prom:example_reasoning1}]
I chose to transfer 26 because as friends, I want to show fairness and kindness, but I also don't want to give away my entire provisional allocation.
\end{lstlisting}

\begin{lstlisting}[language=json,firstnumber=1,caption={\textsc{Example Textual Output of Reasoning (2)}}, label={prom:example_reasoning2}]
I chose to transfer a moderate amount as I want to strike a balance between being fair and not giving away too much of the additional amount I was allocated. Since we will meet after the game, I also consider the social implications of my decision.
\end{lstlisting}

\begin{lstlisting}[language=json,firstnumber=1,caption={\textsc{Example Textual Output of Reasoning (3)}}, label={prom:example_reasoning3}]
As an ESTP personality type, I tend to make decisions based on logic and fairness. Since I don't have any information about the other participant's needs or circumstances, I don't feel inclined to transfer any amount to them. I also don't want to take anything from them, as that would be unfair. Therefore, I choose to keep the additional 47 USD for myself.
\end{lstlisting}

\clearpage
\section{Results}

\subsection{Key Descriptive Statistics}

\subsubsection{Sense of Self (SoS) Trials}

\begin{table}[!h]
    \caption{\textsc{Descriptive statistics for Age (SoS)}} \label{tab:descriptive_age_self}
    \begin{tabular}{lrrrrrrrr}
        \toprule
        Model\_Size       & Count & Mean  & Std   & Min & 25\% & 50\% & 75\% & Max \\
        \midrule
        gemma2\_27b       & 8,271 & 40.04 & 11.79 & 20  & 30   & 40   & 50   & 60  \\
        gemma2\_9b        & 4,582 & 39.66 & 11.91 & 20  & 29   & 40   & 50   & 60  \\
        gpt4o\_2024-08-06 & 9,561 & 39.70 & 11.86 & 20  & 30   & 39   & 50   & 60  \\
        llama3.1\_405b    & 8,997 & 40.01 & 11.76 & 20  & 30   & 40   & 50   & 60  \\
        llama3.1\_70b     & 9,633 & 40.28 & 11.86 & 20  & 30   & 40   & 51   & 60  \\
        llama3.1\_8b      & 4,020 & 39.97 & 11.75 & 20  & 30   & 40   & 50   & 60  \\
        phi3\_14b         & 2,980 & 40.16 & 11.96 & 20  & 29   & 40   & 51   & 60  \\
        phi3\_3.8b        & 773   & 39.26 & 12.08 & 20  & 28   & 39   & 50   & 60  \\
        qwen2.5\_72b      & 5,442 & 40.08 & 11.85 & 20  & 30   & 40   & 50   & 60  \\
        qwen2.5\_7b       & 535   & 39.19 & 12.07 & 20  & 29   & 39   & 50   & 60  \\
        \bottomrule
    \end{tabular}
\end{table}

\clearpage
\begin{table}
    \caption{\textsc{Descriptive statistics for Stake (SoS)}}
    \begin{tabular}{lrrrrrrrr}
        \toprule
        Model\_Size       & Count & Mean  & Std   & Min & 25\%  & 50\% & 75\% & Max \\
        \midrule
        gemma2\_27b       & 8,271 & 55.84 & 26.45 & 10  & 33    & 57   & 78   & 100 \\
        gemma2\_9b        & 4,582 & 54.06 & 27.99 & 10  & 28    & 55   & 79   & 100 \\
        gpt4o\_2024-08-06 & 9,561 & 55.36 & 26.18 & 10  & 33    & 56   & 78   & 100 \\
        llama3.1\_405b    & 8,997 & 54.50 & 26.29 & 10  & 31    & 55   & 77   & 100 \\
        llama3.1\_70b     & 9,633 & 54.67 & 26.13 & 10  & 32    & 55   & 77   & 100 \\
        llama3.1\_8b      & 4,020 & 54.92 & 27.36 & 10  & 31    & 54   & 80   & 100 \\
        phi3\_14b         & 2,980 & 57.74 & 26.78 & 10  & 34    & 62   & 80   & 100 \\
        phi3\_3.8b        & 773   & 49.56 & 27.21 & 10  & 25    & 47   & 70   & 100 \\
        qwen2.5\_72b      & 5,442 & 55.35 & 26.01 & 10  & 32    & 57   & 76   & 100 \\
        qwen2.5\_7b       & 535   & 50.80 & 25.66 & 10  & 29.50 & 52   & 72   & 100 \\
        \bottomrule
    \end{tabular}
\end{table}

\clearpage
\begin{table}
    \caption{\textsc{Descriptive statistics for Temperature (SoS)}}
    \begin{tabular}{lrrrrrrrr}
        \toprule
        Model\_Size       & Count & Mean & Std  & Min & 25\% & 50\% & 75\% & Max \\
        \midrule
        gemma2\_27b       & 8,271 & 0.49 & 0.29 & 0   & 0.24 & 0.49 & 0.74 & 1   \\
        gemma2\_9b        & 4,582 & 0.49 & 0.29 & 0   & 0.25 & 0.49 & 0.73 & 1   \\
        gpt4o\_2024-08-06 & 9,561 & 0.50 & 0.29 & 0   & 0.25 & 0.50 & 0.75 & 1   \\
        llama3.1\_405b    & 8,997 & 0.50 & 0.29 & 0   & 0.25 & 0.50 & 0.75 & 1   \\
        llama3.1\_70b     & 9,633 & 0.50 & 0.29 & 0   & 0.25 & 0.50 & 0.75 & 1   \\
        llama3.1\_8b      & 4,020 & 0.46 & 0.28 & 0   & 0.22 & 0.44 & 0.70 & 1   \\
        phi3\_14b         & 2,980 & 0.45 & 0.29 & 0   & 0.20 & 0.42 & 0.69 & 1   \\
        phi3\_3.8b        & 773   & 0.45 & 0.28 & 0   & 0.20 & 0.41 & 0.68 & 1   \\
        qwen2.5\_72b      & 5,442 & 0.49 & 0.29 & 0   & 0.23 & 0.49 & 0.74 & 1   \\
        qwen2.5\_7b       & 535   & 0.53 & 0.29 & 0   & 0.27 & 0.56 & 0.80 & 1   \\
        \bottomrule
    \end{tabular}
\end{table}

\clearpage
\begin{table}
    \caption{\textsc{Descriptive statistics for Gender (SoS)}}
    \begin{tabular}{lrr}
        \toprule
        Model\_Size       & Male            & Female          \\
        \midrule
        gemma2\_27b       & 4,075 (49.27\%) & 4,196 (50.73\%) \\
        gemma2\_9b        & 2,450 (53.47\%) & 2,132 (46.53\%) \\
        gpt4o\_2024-08-06 & 4,718 (49.35\%) & 4,843 (50.65\%) \\
        llama3.1\_405b    & 4,493 (49.94\%) & 4,504 (50.06\%) \\
        llama3.1\_70b     & 4,767 (49.49\%) & 4,866 (50.51\%) \\
        llama3.1\_8b      & 2,024 (50.35\%) & 1,996 (49.65\%) \\
        phi3\_14b         & 1,495 (50.17\%) & 1,485 (49.83\%) \\
        phi3\_3.8b        & 395 (51.10\%)   & 378 (48.90\%)   \\
        qwen2.5\_72b      & 2,728 (50.13\%) & 2,714 (49.87\%) \\
        qwen2.5\_7b       & 279 (52.15\%)   & 256 (47.85\%)   \\
        \bottomrule
    \end{tabular}
\end{table}

\clearpage
\begin{table}
    \caption{\textsc{Descriptive statistics for marital status (SoS)}}
    \begin{tabular}{lrr}
        \toprule
        Model\_Size       & Currently Married & Not Currently Married \\
        \midrule
        gemma2\_27b       & 4,080 (49.33\%)   & 4,191 (50.67\%)       \\
        gemma2\_9b        & 2,248 (49.06\%)   & 2,334 (50.94\%)       \\
        gpt4o\_2024-08-06 & 4,715 (49.31\%)   & 4,846 (50.69\%)       \\
        llama3.1\_405b    & 4,486 (49.86\%)   & 4,511 (50.14\%)       \\
        llama3.1\_70b     & 4,822 (50.06\%)   & 4,811 (49.94\%)       \\
        llama3.1\_8b      & 1,965 (48.88\%)   & 2,055 (51.12\%)       \\
        phi3\_14b         & 1,504 (50.47\%)   & 1,476 (49.53\%)       \\
        phi3\_3.8b        & 402 (52.01\%)     & 371 (47.99\%)         \\
        qwen2.5\_72b      & 2,698 (49.58\%)   & 2,744 (50.42\%)       \\
        qwen2.5\_7b       & 246 (45.98\%)     & 289 (54.02\%)         \\
        \bottomrule
    \end{tabular}
\end{table}

\clearpage
\begin{table}
    \caption{\textsc{Descriptive statistics for education attainment (SoS)}}
    \begin{threeparttable}
        \begin{tabular}{lrrr}
            \toprule
            Model\_Size       & 0               & 1               & 2               \\
            \midrule
            gemma2\_27b       & 2,789 (33.72\%) & 2,739 (33.12\%) & 2,743 (33.16\%) \\
            gemma2\_9b        & 1,817 (39.66\%) & 1,457 (31.80\%) & 1,308 (28.55\%) \\
            gpt4o\_2024-08-06 & 3,184 (33.30\%) & 3,159 (33.04\%) & 3,218 (33.66\%) \\
            llama3.1\_405b    & 3,072 (34.14\%) & 2,982 (33.14\%) & 2,943 (32.71\%) \\
            llama3.1\_70b     & 3,253 (33.77\%) & 3,132 (32.51\%) & 3,248 (33.72\%) \\
            llama3.1\_8b      & 1,329 (33.06\%) & 1,432 (35.62\%) & 1,259 (31.32\%) \\
            phi3\_14b         & 972 (32.62\%)   & 967 (32.45\%)   & 1,041 (34.93\%) \\
            phi3\_3.8b        & 251 (32.47\%)   & 267 (34.54\%)   & 255 (32.99\%)   \\
            qwen2.5\_72b      & 1,821 (33.46\%) & 1,837 (33.76\%) & 1,784 (32.78\%) \\
            qwen2.5\_7b       & 143 (26.73\%)   & 218 (40.75\%)   & 174 (32.52\%)   \\
            \bottomrule
        \end{tabular}
        \begin{tablenotes}[para,flushleft]
            \footnotesize\textit{Note}: 0 = Less than High School Education; 1 = High School Diploma, but no Four-Year College Degree; 2 = Bachelor's Degree or more.
        \end{tablenotes}
    \end{threeparttable}
\end{table}

\clearpage
\begin{table}
    \caption{\textsc{Descriptive statistics for MBTI Type (SoS)}}
    \begin{threeparttable}
        \begin{tabular}{lrrrr}
            \toprule
            Model\_Size       & (I)ntroversion  & (F)eeling       & i(N)tuition     & (P)erceiving    \\
            \midrule
            gemma2\_27b       & 4,286 (51.82\%) & 3,928 (47.49\%) & 3,989 (48.23\%) & 4,124 (49.86\%) \\
            gemma2\_9b        & 2,608 (56.92\%) & 1,720 (37.54\%) & 2,163 (47.21\%) & 2,365 (51.62\%) \\
            gpt4o\_2024-08-06 & 4,882 (51.06\%) & 4,777 (49.96\%) & 4,842 (50.64\%) & 4,752 (49.70\%) \\
            llama3.1\_405b    & 4,545 (50.52\%) & 4,317 (47.98\%) & 4,406 (48.97\%) & 4,442 (49.37\%) \\
            llama3.1\_70b     & 4,832 (50.16\%) & 4,688 (48.67\%) & 4,797 (49.80\%) & 4,821 (50.05\%) \\
            llama3.1\_8b      & 2,101 (52.26\%) & 2,019 (50.22\%) & 1,978 (49.20\%) & 2,026 (50.40\%) \\
            phi3\_14b         & 1,483 (49.77\%) & 1,505 (50.50\%) & 1,527 (51.24\%) & 1,460 (48.99\%) \\
            phi3\_3.8b        & 356 (46.05\%)   & 386 (49.94\%)   & 420 (54.33\%)   & 389 (50.32\%)   \\
            qwen2.5\_72b      & 2,714 (49.87\%) & 2,679 (49.23\%) & 2,723 (50.04\%) & 2,713 (49.85\%) \\
            qwen2.5\_7b       & 218 (40.75\%)   & 306 (57.20\%)   & 306 (57.20\%)   & 284 (53.08\%)   \\
            \bottomrule
        \end{tabular}
        \begin{tablenotes}[para,flushleft]
            \footnotesize\textit{Note}: Proportions are by MBTI types. For example, 51.82\% of the participants in the gemma2\_27b model are Introversion, which means that 48.18\% are Extraversion.
        \end{tablenotes}
    \end{threeparttable}
\end{table}

\clearpage
\begin{table}
    \caption{\textsc{Descriptive statistics for Amount Transfer (SoS)}}
    \begin{tabular}{lrrrrrrrr}
        \toprule
        Model\_Size       & Count & Mean  & Std   & Min & 25\% & 50\%  & 75\%  & Max \\
        \midrule
        gemma2\_27b       & 8,271 & 11.40 & 15.52 & 0   & 0    & 0     & 21    & 98  \\
        gemma2\_9b        & 4,582 & 7.68  & 13.25 & 0   & 0    & 0     & 12    & 60  \\
        gpt4o\_2024-08-06 & 9,561 & 26.05 & 16.06 & -20 & 13   & 26    & 39    & 99  \\
        llama3.1\_405b    & 8,997 & 17.45 & 16.69 & -23 & 0    & 13    & 32.50 & 50  \\
        llama3.1\_70b     & 9,633 & 10.63 & 14.66 & 0   & 0    & 0     & 19    & 89  \\
        llama3.1\_8b      & 4,020 & 4.44  & 11.22 & -85 & 0    & 0     & 0     & 83  \\
        phi3\_14b         & 2,980 & 21.79 & 16.79 & -10 & 6    & 21    & 36    & 50  \\
        phi3\_3.8b        & 773   & 24.58 & 13.74 & 0   & 12   & 23.50 & 35    & 50  \\
        qwen2.5\_72b      & 5,442 & 21.98 & 14.80 & 0   & 10   & 21    & 34    & 50  \\
        qwen2.5\_7b       & 535   & 24.87 & 13.37 & 0   & 14   & 26    & 36    & 50  \\
        \bottomrule
    \end{tabular}
\end{table}

\clearpage
\subsubsection{Theory of Mind (ToM) Trials}

\begin{table}[htbp]
    \caption{\textsc{Descriptive statistics for Age (ToM)}}
    \begin{tabular}{lrrrrrrrr}
        \toprule
        Model\_Size       & Count & Mean  & Std   & Min & 25\%  & 50\% & 75\% & Max \\
        \midrule
        gemma2\_27b       & 2,341 & 39.34 & 11.70 & 20  & 29    & 39   & 49   & 60  \\
        gemma2\_9b        & 594   & 39.41 & 12.16 & 20  & 28.25 & 39   & 50   & 60  \\
        gpt4o\_2024-08-06 & 9,692 & 40.03 & 11.90 & 20  & 30    & 40   & 51   & 60  \\
        llama3.1\_405b    & 8,970 & 40.25 & 11.89 & 20  & 30    & 40   & 51   & 60  \\
        llama3.1\_70b     & 9,227 & 39.78 & 11.85 & 20  & 30    & 40   & 50   & 60  \\
        llama3.1\_8b      & 4,540 & 40    & 11.84 & 20  & 30    & 40   & 50   & 60  \\
        phi3\_14b         & 2,693 & 39.61 & 11.95 & 20  & 29    & 39   & 50   & 60  \\
        phi3\_3.8b        & 1,551 & 39.38 & 11.49 & 20  & 30    & 39   & 49   & 60  \\
        qwen2.5\_72b      & 5,184 & 40.52 & 11.73 & 20  & 30    & 41   & 51   & 60  \\
        qwen2.5\_7b       & 2,018 & 39.42 & 11.64 & 20  & 29    & 39   & 50   & 60  \\
        \bottomrule
    \end{tabular}
\end{table}

\clearpage
\begin{table}
    \caption{\textsc{Descriptive statistics for Stake (ToM)}}
    \begin{tabular}{lrrrrrrrr}
        \toprule
        Model\_Size       & Count & Mean  & Std   & Min & 25\%  & 50\% & 75\% & Max \\
        \midrule
        gemma2\_27b       & 2,341 & 50.78 & 28.75 & 10  & 24    & 48   & 76   & 100 \\
        gemma2\_9b        & 594   & 38.43 & 23.13 & 10  & 18.25 & 32   & 52   & 100 \\
        gpt4o\_2024-08-06 & 9,692 & 55.82 & 26.24 & 10  & 33    & 56   & 79   & 100 \\
        llama3.1\_405b    & 8,970 & 54.15 & 25.91 & 10  & 32    & 54   & 76   & 100 \\
        llama3.1\_70b     & 9,227 & 54.01 & 26.09 & 10  & 32    & 53   & 76   & 100 \\
        llama3.1\_8b      & 4,540 & 55.55 & 26.76 & 10  & 32    & 57   & 79   & 100 \\
        phi3\_14b         & 2,693 & 55.38 & 27.30 & 10  & 32    & 56   & 80   & 100 \\
        phi3\_3.8b        & 1,551 & 50.78 & 29.15 & 10  & 25    & 43   & 80   & 100 \\
        qwen2.5\_72b      & 5,184 & 54    & 26.72 & 10  & 30    & 54   & 77   & 100 \\
        qwen2.5\_7b       & 2,018 & 55.03 & 25.03 & 10  & 36    & 54   & 74   & 100 \\
        \bottomrule
    \end{tabular}
\end{table}

\clearpage
\begin{table}
    \caption{\textsc{Descriptive statistics for Temperature (ToM)}}
    \begin{tabular}{lrrrrrrrr}
        \toprule
        Model\_Size       & Count & Mean & Std  & Min & 25\% & 50\% & 75\% & Max \\
        \midrule
        gemma2\_27b       & 2,341 & 0.51 & 0.29 & 0   & 0.26 & 0.51 & 0.77 & 1   \\
        gemma2\_9b        & 594   & 0.53 & 0.29 & 0   & 0.28 & 0.55 & 0.78 & 1   \\
        gpt4o\_2024-08-06 & 9,692 & 0.50 & 0.29 & 0   & 0.25 & 0.51 & 0.75 & 1   \\
        llama3.1\_405b    & 8,970 & 0.50 & 0.29 & 0   & 0.25 & 0.50 & 0.75 & 1   \\
        llama3.1\_70b     & 9,227 & 0.49 & 0.29 & 0   & 0.25 & 0.49 & 0.74 & 1   \\
        llama3.1\_8b      & 4,540 & 0.46 & 0.28 & 0   & 0.23 & 0.45 & 0.68 & 1   \\
        phi3\_14b         & 2,693 & 0.45 & 0.28 & 0   & 0.21 & 0.43 & 0.68 & 1   \\
        phi3\_3.8b        & 1,551 & 0.47 & 0.28 & 0   & 0.23 & 0.46 & 0.72 & 1   \\
        qwen2.5\_72b      & 5,184 & 0.50 & 0.29 & 0   & 0.25 & 0.50 & 0.75 & 1   \\
        qwen2.5\_7b       & 2,018 & 0.47 & 0.28 & 0   & 0.23 & 0.46 & 0.70 & 1   \\
        \bottomrule
    \end{tabular}
\end{table}

\clearpage
\begin{table}
    \caption{\textsc{Descriptive statistics for Gender (ToM)}}
    \begin{tabular}{lrr}
        \toprule
        Model\_Size       & Male            & Female          \\
        \midrule
        gemma2\_27b       & 1,292 (55.19\%) & 1,049 (44.81\%) \\
        gemma2\_9b        & 328 (55.22\%)   & 266 (44.78\%)   \\
        gpt4o\_2024-08-06 & 4,820 (49.73\%) & 4,872 (50.27\%) \\
        llama3.1\_405b    & 4,512 (50.30\%) & 4,458 (49.70\%) \\
        llama3.1\_70b     & 4,637 (50.25\%) & 4,590 (49.75\%) \\
        llama3.1\_8b      & 2,279 (50.20\%) & 2,261 (49.80\%) \\
        phi3\_14b         & 1,282 (47.60\%) & 1,411 (52.40\%) \\
        phi3\_3.8b        & 762 (49.13\%)   & 789 (50.87\%)   \\
        qwen2.5\_72b      & 2,532 (48.84\%) & 2,652 (51.16\%) \\
        qwen2.5\_7b       & 1,014 (50.25\%) & 1,004 (49.75\%) \\
        \bottomrule
    \end{tabular}
\end{table}

\clearpage
\begin{table}
    \caption{\textsc{Descriptive statistics for marital status (ToM)}}
    \begin{tabular}{lrr}
        \toprule
        Model\_Size       & Currently Married & Not Currently Married \\
        \midrule
        gemma2\_27b       & 1,050 (44.85\%)   & 1,291 (55.15\%)       \\
        gemma2\_9b        & 289 (48.65\%)     & 305 (51.35\%)         \\
        gpt4o\_2024-08-06 & 4,756 (49.07\%)   & 4,936 (50.93\%)       \\
        llama3.1\_405b    & 4,491 (50.07\%)   & 4,479 (49.93\%)       \\
        llama3.1\_70b     & 4,681 (50.73\%)   & 4,546 (49.27\%)       \\
        llama3.1\_8b      & 2,267 (49.93\%)   & 2,273 (50.07\%)       \\
        phi3\_14b         & 1,381 (51.28\%)   & 1,312 (48.72\%)       \\
        phi3\_3.8b        & 762 (49.13\%)     & 789 (50.87\%)         \\
        qwen2.5\_72b      & 2,590 (49.96\%)   & 2,594 (50.04\%)       \\
        qwen2.5\_7b       & 1,023 (50.69\%)   & 995 (49.31\%)         \\
        \bottomrule
    \end{tabular}
\end{table}

\clearpage
\begin{table}
    \caption{\textsc{Descriptive statistics for education attainment (ToM)}}
    \begin{threeparttable}
        \begin{tabular}{lrrr}
            \toprule
            Model\_Size       & 0               & 1               & 2               \\
            \midrule
            gemma2\_27b       & 962 (41.09\%)   & 628 (26.83\%)   & 751 (32.08\%)   \\
            gemma2\_9b        & 257 (43.27\%)   & 153 (25.76\%)   & 184 (30.98\%)   \\
            gpt4o\_2024-08-06 & 3,277 (33.81\%) & 3,208 (33.10\%) & 3,207 (33.09\%) \\
            llama3.1\_405b    & 2,991 (33.34\%) & 3,073 (34.26\%) & 2,906 (32.40\%) \\
            llama3.1\_70b     & 3,055 (33.11\%) & 3,054 (33.10\%) & 3,118 (33.79\%) \\
            llama3.1\_8b      & 1,463 (32.22\%) & 1,657 (36.50\%) & 1,420 (31.28\%) \\
            phi3\_14b         & 752 (27.92\%)   & 912 (33.87\%)   & 1,029 (38.21\%) \\
            phi3\_3.8b        & 347 (22.37\%)   & 554 (35.72\%)   & 650 (41.91\%)   \\
            qwen2.5\_72b      & 1,697 (32.74\%) & 1,683 (32.47\%) & 1,804 (34.80\%) \\
            qwen2.5\_7b       & 618 (30.62\%)   & 696 (34.49\%)   & 704 (34.89\%)   \\
            \bottomrule
        \end{tabular}
        \begin{tablenotes}[para,flushleft]
            \footnotesize\textit{Note}: 0 = Less than High School Education; 1 = High School Diploma, but no Four-Year College Degree; 2 = Bachelor's Degree or more.
        \end{tablenotes}
    \end{threeparttable}
\end{table}

\clearpage
\begin{table}
    \caption{\textsc{Descriptive statistics for MBTI Type (ToM)}}
    \label{tab:mbti_tom}
    \begin{threeparttable}
        \begin{tabular}{lrrrr}
            \toprule
            Model\_Size       & (I)ntroversion  & (F)eeling       & i(N)tuition     & (P)erceiving    \\
            \midrule
            gemma2\_27b       & 1,205 (51.47\%) & 867 (37.04\%)   & 1,158 (49.47\%) & 1,127 (48.14\%) \\
            gemma2\_9b        & 291 (48.99\%)   & 261 (43.94\%)   & 309 (52.02\%)   & 281 (47.31\%)   \\
            gpt4o\_2024-08-06 & 4,916 (50.72\%) & 4,799 (49.52\%) & 4,853 (50.07\%) & 4,824 (49.77\%) \\
            llama3.1\_405b    & 4,532 (50.52\%) & 4,404 (49.10\%) & 4,408 (49.14\%) & 4,460 (49.72\%) \\
            llama3.1\_70b     & 4,552 (49.33\%) & 4,649 (50.38\%) & 4,605 (49.91\%) & 4,570 (49.53\%) \\
            llama3.1\_8b      & 2,286 (50.35\%) & 2,191 (48.26\%) & 2,216 (48.81\%) & 2,284 (50.31\%) \\
            phi3\_14b         & 1,252 (46.49\%) & 1,432 (53.17\%) & 1,407 (52.25\%) & 1,248 (46.34\%) \\
            phi3\_3.8b        & 668 (43.07\%)   & 784 (50.55\%)   & 823 (53.06\%)   & 773 (49.84\%)   \\
            qwen2.5\_72b      & 2,580 (49.77\%) & 2,600 (50.15\%) & 2,700 (52.08\%) & 2,462 (47.49\%) \\
            qwen2.5\_7b       & 959 (47.52\%)   & 1,105 (54.76\%) & 1,050 (52.03\%) & 1,021 (50.59\%) \\
            \bottomrule
        \end{tabular}
        \begin{tablenotes}[para,flushleft]
            \footnotesize\textit{Note}: Proportions are by MBTI types. For example, 51.82\% of the participants in the gemma2\_27b model are Introversion, which means that 48.18\% are Extraversion.
        \end{tablenotes}
    \end{threeparttable}
\end{table}

\clearpage
\begin{table}
    \caption{\textsc{Descriptive statistics for Amount Transfer (ToM)}}
    \label{tab:amount_transfer_tom}
    \begin{tabular}{lrrrrrrrr}
        \toprule
        Model\_Size       & Count & Mean  & Std   & Min & 25\%  & 50\%  & 75\% & Max \\
        \midrule
        gemma2\_27b       & 2,341 & 14.98 & 16.50 & 0   & 0     & 8     & 28   & 50  \\
        gemma2\_9b        & 594   & 13.98 & 13.28 & 0   & 0     & 12    & 22   & 50  \\
        gpt4o\_2024-08-06 & 9,692 & 24.45 & 12.98 & 0   & 14    & 21    & 35   & 91  \\
        llama3.1\_405b    & 8,970 & 19.42 & 13.80 & 0   & 10    & 16    & 30   & 70  \\
        llama3.1\_70b     & 9,227 & 14.68 & 13.90 & -20 & 5     & 10    & 21   & 91  \\
        llama3.1\_8b      & 4,540 & 2.66  & 8.32  & -10 & 0     & 0     & 0    & 50  \\
        phi3\_14b         & 2,693 & 27.13 & 14.61 & -20 & 15    & 27    & 40   & 98  \\
        phi3\_3.8b        & 1,551 & 25.37 & 14.62 & 0   & 12.50 & 21.50 & 40   & 64  \\
        qwen2.5\_72b      & 5,184 & 21.18 & 13.65 & 0   & 10    & 20    & 31   & 50  \\
        qwen2.5\_7b       & 2,018 & 27.49 & 12.73 & -10 & 17.12 & 27    & 37   & 79  \\
        \bottomrule
    \end{tabular}
\end{table}

\clearpage
\subsection{Experiment Results of Theory of Mind (ToM) Trials} \label{sec:tom_results}

\begin{table}[!h]
    \caption{\textsc{Model Performance: Instruction following and math reasoning (ToM)}} \label{tab:model_performance_tom}
    \small
    \begin{threeparttable}
        \begin{tabular}{>{\raggedleft\arraybackslash}m{1cm}m{3cm}>{\raggedleft\arraybackslash}m{2cm}>{\raggedleft\arraybackslash}m{2cm}>{\raggedleft\arraybackslash}m{2cm}>{\raggedleft\arraybackslash}m{2cm}}
            \hline\hline
               & {Model\_Size}     & {\#Simulation Trials} & {\#Correct JSON Format} & {\#Logically Correct Trials} & {\%Logically Correct Trials} \\
            \hline
            1  & gpt4o\_2024-08-06 & 10,000                & 10,000                  & 9,692                        & 96.92                        \\
            2  & llama3.1\_70b     & 10,000                & 9,998                   & 9,227                        & 92.29                        \\
            3  & llama3.1\_405b    & 10,000                & 9,979                   & 8,970                        & 89.89                        \\
            4  & qwen2.5\_72b      & 10,000                & 10,000                  & 5,184                        & 51.84                        \\
            5  & llama3.1\_8b      & 10,000                & 9,986                   & 4,540                        & 45.46                        \\
            6  & phi3\_14b         & 10,000                & 9,911                   & 2,693                        & 27.17                        \\
            7  & gemma2\_27b       & 10,000                & 9,994                   & 2,341                        & 23.42                        \\
            8  & qwen2.5\_7b       & 10,000                & 9,945                   & 2,018                        & 20.29                        \\
            9  & phi3\_3.8b        & 10,000                & 9,783                   & 1,551                        & 15.85                        \\
            10 & gemma2\_9b        & 10,000                & 9,473                   & 594                          & 6.27                         \\
            \hline\hline
        \end{tabular}
        \begin{tablenotes}[para,flushleft]
            \footnotesize\textit{Note}: ``\#Correct JSON Format'' indicates the number of responses in correct JSON format, suggesting a model's ability of instruction following. ``\#Logically Correct Trials'' and ``\%Logically Correct Trials'' indicate the number and corresponding percentage of responses that are logically correct, suggesting a model's ability of math reasoning. Results of the Sense of Self trials are in Table \ref{tab:model_performance_self}.
        \end{tablenotes}
    \end{threeparttable}
\end{table}

\clearpage
\begin{sidewaysfigure}[!htbp]
    \centering
    \caption{\textsc{Giving rate by model family and size (ToM)}}
    \label{fig:giving_rate_tom}
    \includegraphics[width=\textwidth]{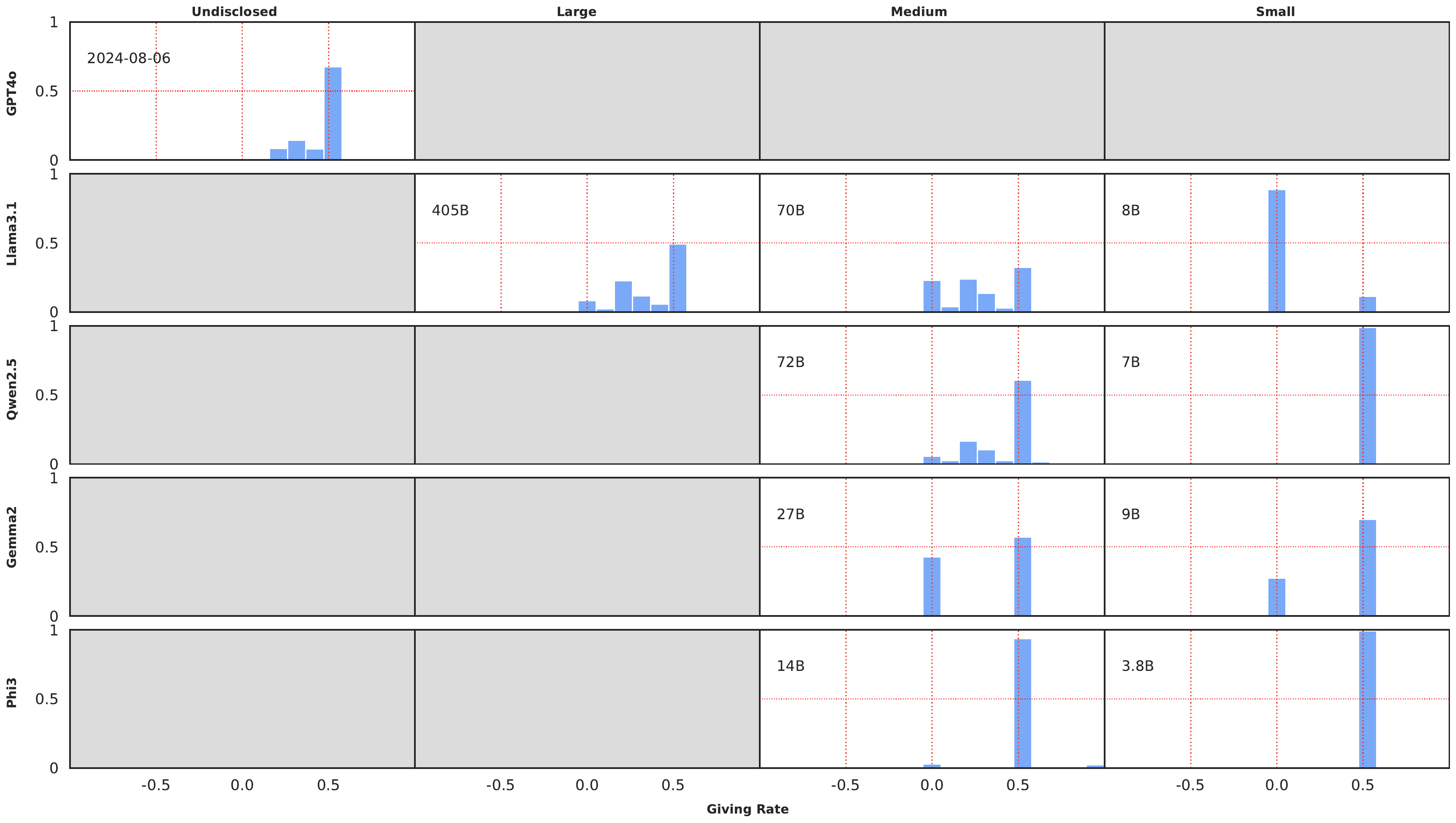}
    \vspace{0.5em} 
    \begin{minipage}{\textwidth}
        \footnotesize\textit{Note}: Vertical red dashed lines indicate giving rates at -0.5, 0, and 0.5, respectively; horizontal red dashed lines indicate 50\% of total observations. The giving rate is calculated as the percentage of the amount transferred by the dictator to the recipient out of the total stake. Results of the Sense of Self trials are in Figure \ref{fig:giving_rate_self}.
    \end{minipage}
\end{sidewaysfigure}

\begin{figure}[!htbp]
    \centering
    \caption{\textsc{Predicting generosity: Demographics and LLM temperature (ToM)}}
    \label{fig:demographics_tom}
    \includegraphics[width=1\textwidth]{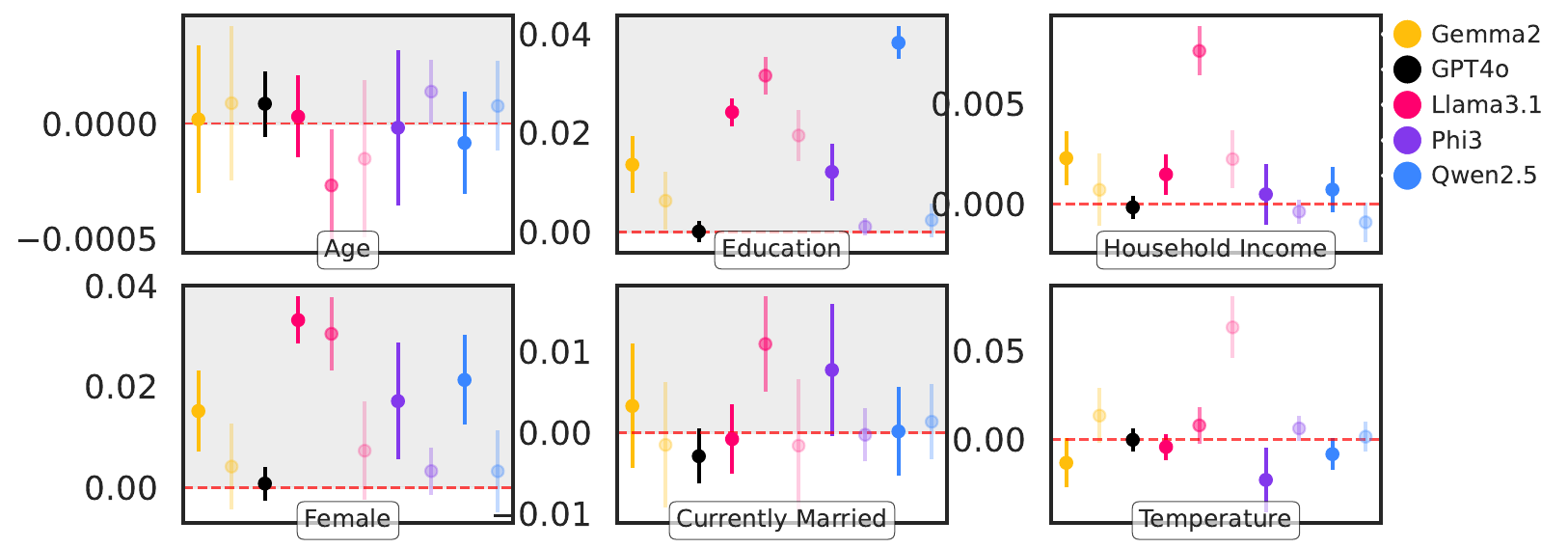}
    \begin{minipage}[t]{\linewidth}
        \footnotesize\textit{Note}: The coefficients (showing 95\% confidence intervals) are from a linear regression model using money transferred in the dictator game as the dependent variable. Deep colors represent larger models, and light colors represent smaller models within the same LLM family. The shaded areas indicate expected directions of impact based on human studies (Appendix \ref{sec:human_baseline_factors}). Results of the Sense of Self trials are in Figure \ref{fig:demographics_self}.
    \end{minipage}
\end{figure}

\begin{figure}[!htbp]
    \centering
    \caption{\textsc{Predicting generosity: Myers–Briggs Type Indicator (ToM)}}
    \label{fig:mbti_tom}
    \includegraphics[width=0.8\textwidth]{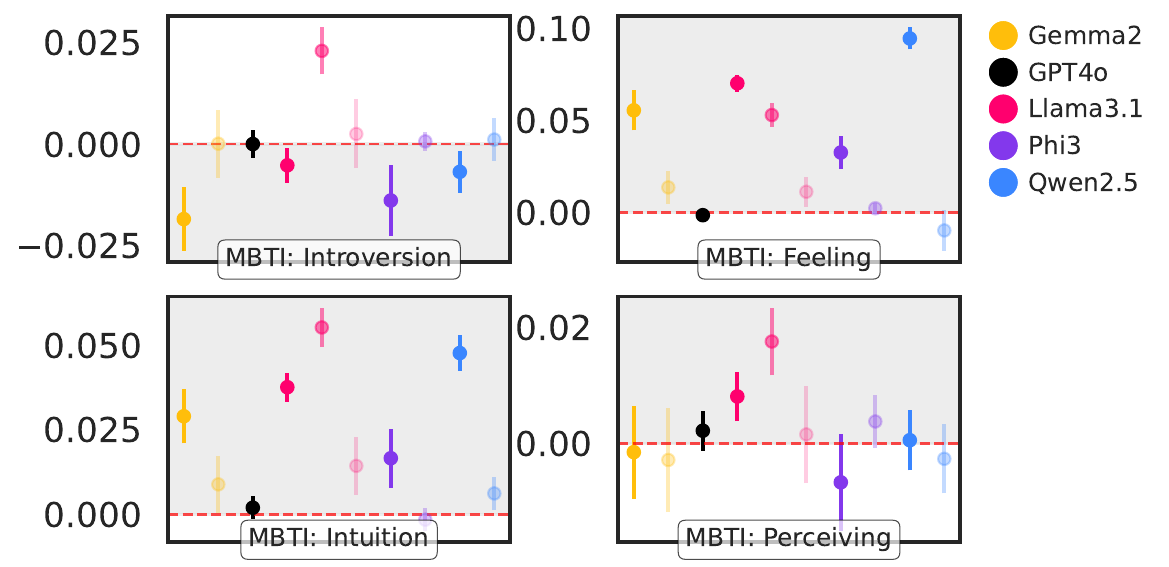}
    \begin{minipage}[t]{0.8\linewidth}
        \footnotesize\textit{Note}: The coefficients (showing 95\% confidence intervals) are from a linear regression model using money transferred in the dictator game as the dependent variable. Deep colors represent larger models, and light colors represent smaller models within the same LLM family. The shaded areas indicate expected directions of impact based on human studies (Appendix \ref{sec:human_baseline_factors}). Results of the Sense of Self trials are in Figure \ref{fig:mbti_self}.
    \end{minipage}
\end{figure}

\begin{figure}[!htbp]
    \centering
    \caption{\textsc{Predicting generosity: Framing of experiment (ToM)}}
    \label{fig:framing_tom}
    \includegraphics[width=0.8\textwidth]{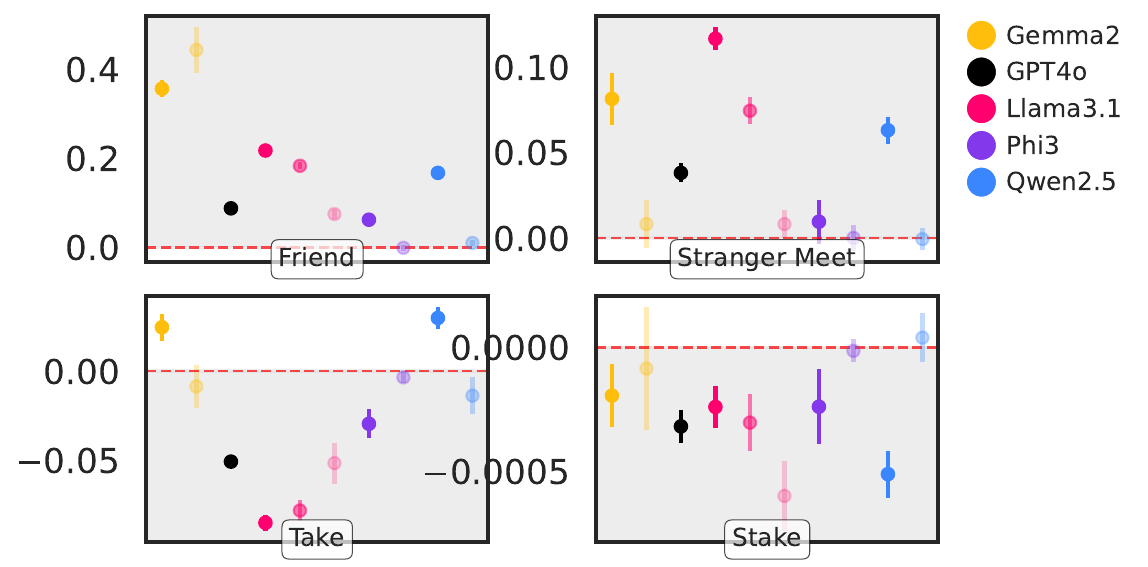}
    \begin{minipage}[t]{\linewidth}
        \footnotesize\textit{Note}: The coefficients (showing 95\% confidence intervals) are from a linear regression model using money transferred in the dictator game as the dependent variable. Deep colors represent larger models, and light colors represent smaller models within the same LLM family. The shaded areas indicate expected directions of impact based on human studies (Appendix \ref{sec:human_baseline_factors}). The ``Stranger'' framing is the reference group for ``Friend'' and ``Stranger Meet.'' The ``Give'' framing is the reference group for ``Take.'' Results of the Sense of Self trials are in Figure \ref{fig:framing_self}.
    \end{minipage}
\end{figure}

\begin{sidewaysfigure}[!htbp]
    \centering
    \caption{\textsc{Predicting generosity: Psychological process (TOM)}}
    \label{fig:liwc_tom}
    \begin{subfigure}{1\textwidth}
        \centering
        \caption{LIWC Categories Effectively Predicting Compassion Controlling for Empathy}
        \includegraphics[width=1\textwidth]{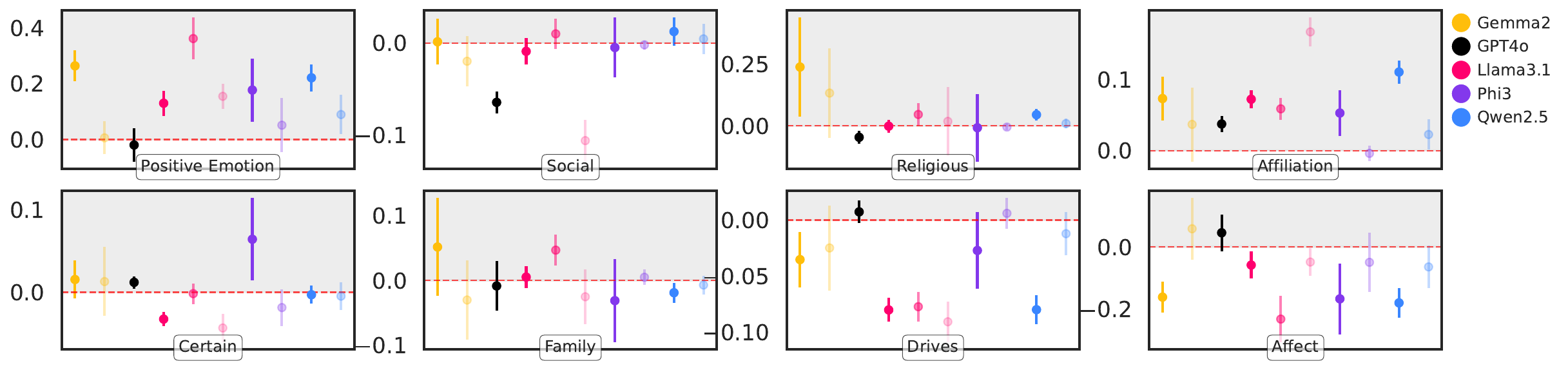}
        \label{fig:compassion_tom}
    \end{subfigure}
    \begin{subfigure}{1\textwidth}
        \centering
        \caption{LIWC Categories Effectively Predicting Empathy Controlling for Compassion}
        \includegraphics[width=1\textwidth]{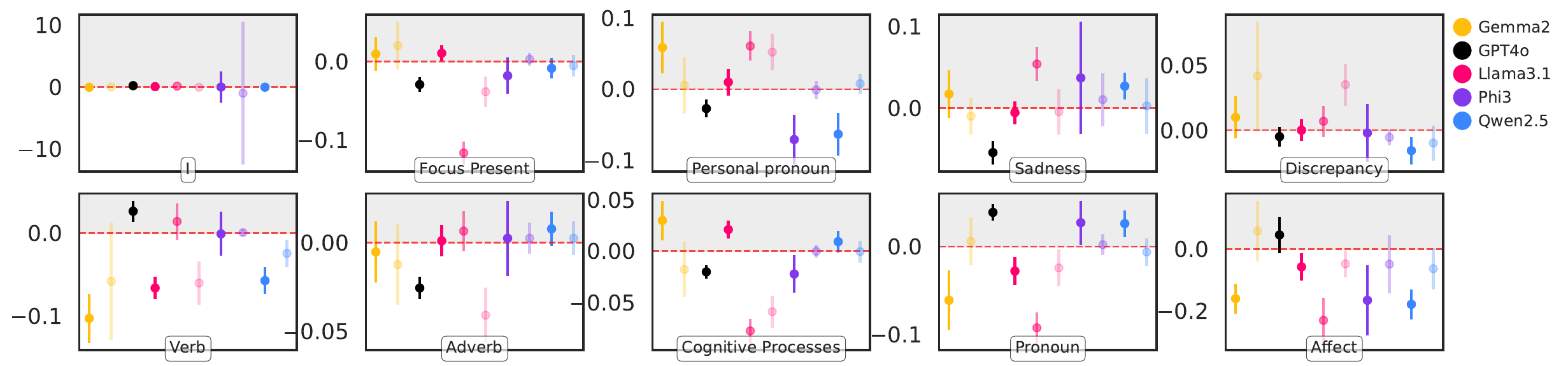}
        \label{fig:empathy_tom}
    \end{subfigure}
    \begin{minipage}[t]{\linewidth}
        \footnotesize\textit{Note}: The coefficients (showing 95\% confidence intervals) are from a linear regression model using money transferred in the dictator game as the dependent variable. Deep colors represent larger models, and light colors represent smaller models within the same LLM family. The shaded areas indicate expected directions of impact based on human studies (Appendix \ref{sec:human_baseline_factors}). LIWC categories are selected for analysis according to \textcite{YadenCharacterizingEmpathyCompassion2024}. ``She/He'' and ``Male'' categories for Compassion are excluded due to limited number of observations. LIWC = Linguistic Inquiry and Word Count \autocite{TausczikPsychologicalMeaningWords2010}. Results of the Sense of Self trials are in Figure \ref{fig:liwc_self}.
    \end{minipage}
\end{sidewaysfigure}

\begin{sidewaystable}[]
    \caption{\textsc{LLM Agent's Alignment with Humans in Dictator Games (ToM)}} \label{tab:alignment_tom}
    \begin{threeparttable}
        \begin{tabular}{rrccccccccccc}
            \hline\hline
               &                                                 & (1)    & (2)    & (3)    & (4)    & (5)    & (6)    & (7)    & (8)    & (9)    & (10)   & Total \cmark \\
               &                                                 & G27B   & G9B    & GPT4o  & L405B  & L70B   & L8B    & P14B   & P3.8B  & Q72B   & Q7B    & (by row)     \\
            \hline
               & \multicolumn{1}{l}{\textit{Demographics}}       &        &        &        &        &        &        &        &        &        &        &              \\
            1  & Age                                             & n.s.   & n.s.   & n.s.   & n.s.   & \xmark & n.s.   & n.s.   & \cmark & n.s.   & n.s.   & 1            \\
            2  & Education                                       & \cmark & \cmark & n.s.   & \cmark & \cmark & \cmark & \cmark & n.s.   & \cmark & n.s.   & 7            \\
            3  & H. Income                                       & pos.   & n.s.   & n.s.   & pos.   & pos.   & pos.   & n.s.   & n.s.   & n.s.   & n.s.   & --           \\
            4  & Female                                          & \cmark & n.s.   & n.s.   & \cmark & \cmark & n.s.   & \cmark & n.s.   & \cmark & n.s.   & 5            \\
            5  & Married                                         & n.s.   & n.s.   & n.s.   & n.s.   & \cmark & n.s.   & n.s.   & n.s.   & n.s.   & n.s.   & 1            \\
            6  & Temperature                                     & n.s.   & n.s.   & n.s.   & n.s.   & n.s.   & pos.   & neg.   & n.s.   & n.s.   & n.s.   & --           \\
               & \multicolumn{1}{l}{\textit{MBTI}}               &        &        &        &        &        &        &        &        &        &        &              \\
            7  & Introversion                                    & \cmark & n.s.   & n.s.   & \cmark & \xmark & n.s.   & \cmark & n.s.   & \cmark & n.s.   & 4            \\
            8  & Feeling                                         & \cmark & \cmark & n.s.   & \cmark & \cmark & \cmark & \cmark & n.s.   & \cmark & n.s.   & 7            \\
            9  & Intuition                                       & \cmark & \cmark & n.s.   & \cmark & \cmark & \cmark & \cmark & n.s.   & \cmark & \cmark & 8            \\
            10 & Perceiving                                      & n.s.   & n.s.   & n.s.   & \cmark & \cmark & n.s.   & n.s.   & n.s.   & n.s.   & n.s.   & 2            \\
               & \multicolumn{1}{l}{\textit{Experiment Framing}} &        &        &        &        &        &        &        &        &        &        &              \\
            11 & Friend                                          & \cmark & \cmark & \cmark & \cmark & \cmark & \cmark & \cmark & n.s.   & \cmark & \cmark & 9            \\
            12 & Stranger Meet                                   & \cmark & n.s.   & \cmark & \cmark & \cmark & n.s.   & n.s.   & n.s.   & \cmark & n.s.   & 5            \\
            13 & Take                                            & \xmark & n.s.   & \cmark & \cmark & \cmark & \cmark & \cmark & \cmark & \xmark & \cmark & 7            \\
            14 & Stake                                           & \cmark & n.s.   & \cmark & \cmark & \cmark & \cmark & \cmark & n.s.   & \cmark & n.s.   & 7            \\
            \hline
               & Total \cmark                                    & 8      & 4      & 4      & 10     & 10     & 6      & 8      & 2      & 8      & 3      & 63           \\
            \hline\hline
        \end{tabular}
        \begin{tablenotes}[para,flushleft]
            \footnotesize\textit{Note}: \cmark = Aligning with human studies; \xmark = Not aligning with human studies; n.s. = Not significant; pos. = Positive; neg. = Negative. ``--'' indicates the lack of consensus from human studies, showing directions of coefficients but not alignments for these variables. The expected directions of impact based on human studies are reviewed in Appendix \ref{sec:human_baseline_factors}. Results for the Sense of Self trials are in Table \ref{tab:alignment_self}.
        \end{tablenotes}
    \end{threeparttable}
\end{sidewaystable}

\begin{sidewaystable}[]
    \caption{\textsc{LLM Agent's Alignment with Humans in Dictator Games: Compassion (ToM)}} \label{tab:compassion_tom}
    \begin{threeparttable}
        \begin{tabular}{rrccccccccccc}
            \hline\hline
              &              & (1)    & (2)  & (3)    & (4)    & (5)    & (6)    & (7)    & (8)   & (9)    & (10)   & Total \cmark \\
              &              & G27B   & G9B  & GPT4o  & L405B  & L70B   & L8B    & P14B   & P3.8B & Q72B   & Q7B    & (by row)     \\
            \hline
            1 & Pos. Emotion & \cmark & n.s. & n.s.   & \cmark & \cmark & \cmark & \cmark & n.s.  & \cmark & \cmark & 7            \\
            2 & Social       & n.s.   & n.s. & \xmark & n.s.   & n.s.   & \xmark & n.s.   & n.s.  & n.s.   & n.s.   & 0            \\
            3 & Religious    & \cmark & n.s. & \xmark & n.s.   & n.s.   & n.s.   & n.s.   & n.s.  & \cmark & n.s.   & 2            \\
            4 & Affiliation  & \cmark & n.s. & \cmark & \cmark & \cmark & \cmark & \cmark & n.s.  & \cmark & \cmark & 8            \\
            5 & Certain      & n.s.   & n.s. & \cmark & \xmark & n.s.   & \xmark & \cmark & n.s.  & n.s.   & n.s.   & 2            \\
            6 & Family       & n.s.   & n.s. & n.s.   & n.s.   & \cmark & n.s.   & n.s.   & n.s.  & \xmark & n.s.   & 1            \\
            7 & Drives       & \xmark & n.s. & n.s.   & \xmark & \xmark & \xmark & n.s.   & n.s.  & \xmark & n.s.   & 0            \\
            8 & Affect       & \xmark & n.s. & n.s.   & \xmark & \xmark & \xmark & \xmark & n.s.  & \xmark & n.s.   & 0            \\
            \hline
              & Total \cmark & 3      & 0    & 2      & 2      & 3      & 2      & 3      & 0     & 3      & 2      & 20           \\
            \hline\hline
        \end{tabular}
        \begin{tablenotes}[para,flushleft]
            \footnotesize\textit{Note}: \cmark = Aligning with human studies; \xmark = Not aligning with human studies; n.s. = Not significant; pos. = Positive; neg. = Negative. ``--'' indicates the lack of consensus from human studies, showing directions of coefficients but not alignments for these variables. The expected directions of impact based on human studies are reviewed in Appendix \ref{sec:human_baseline_factors}. Results for the Sense of Self trials are in Table \ref{tab:compassion_self}.
        \end{tablenotes}
    \end{threeparttable}
\end{sidewaystable}

\begin{sidewaystable}[]
    \caption{\textsc{LLM Agent's Alignment with Humans in Dictator Games: Empathy (ToM)}} \label{tab:empathy_tom}
    \begin{threeparttable}
        \begin{tabular}{rrccccccccccc}
            \hline\hline
               &                     & (1)    & (2)  & (3)    & (4)    & (5)    & (6)    & (7)    & (8)   & (9)    & (10)   & Total \cmark \\
               &                     & G27B   & G9B  & GPT4o  & L405B  & L70B   & L8B    & P14B   & P3.8B & Q72B   & Q7B    & (by row)     \\
            \hline
            1  & I                   & n.s.   & n.s. & n.s.   & n.s.   & n.s.   & n.s.   & n.s.   & n.s.  & n.s.   & n.s.   & 0            \\
            2  & Focus Present       & n.s.   & n.s. & \xmark & \cmark & \xmark & \xmark & n.s.   & n.s.  & n.s.   & n.s.   & 1            \\
            3  & Personal Pronoun    & \cmark & n.s. & \xmark & n.s.   & \cmark & \cmark & \xmark & n.s.  & \xmark & n.s.   & 3            \\
            4  & Sadness             & n.s.   & n.s. & \xmark & n.s.   & \cmark & n.s.   & n.s.   & n.s.  & \cmark & n.s.   & 2            \\
            5  & Discrepancy         & n.s.   & n.s. & n.s.   & n.s.   & n.s.   & \cmark & n.s.   & n.s.  & \xmark & n.s.   & 1            \\
            6  & Verb                & \xmark & n.s. & \cmark & \xmark & n.s.   & \xmark & n.s.   & n.s.  & \xmark & \xmark & 1            \\
            7  & Adverb              & n.s.   & n.s. & \xmark & n.s.   & n.s.   & \xmark & n.s.   & n.s.  & n.s.   & n.s.   & 0            \\
            8  & Cognitive Processes & \cmark & n.s. & \xmark & \cmark & \xmark & \xmark & \xmark & n.s.  & n.s.   & n.s.   & 1            \\
            9  & Pronoun             & \xmark & n.s. & \cmark & \xmark & \xmark & \xmark & \cmark & n.s.  & \cmark & n.s.   & 3            \\
            10 & Affect              & \xmark & n.s. & n.s.   & \xmark & \xmark & \xmark & \xmark & n.s.  & \xmark & n.s.   & 0            \\
            \hline
               & Total \cmark        & 2      & 0    & 2      & 2      & 2      & 2      & 1      & 0     & 2      & 0      & 13           \\
            \hline\hline
        \end{tabular}
        \begin{tablenotes}[para,flushleft]
            \footnotesize\textit{Note}: \cmark = Aligning with human studies; \xmark = Not aligning with human studies; n.s. = Not significant; pos. = Positive; neg. = Negative. ``--'' indicates the lack of consensus from human studies, showing directions of coefficients but not alignments for these variables. The expected directions of impact based on human studies are reviewed in Appendix \ref{sec:human_baseline_factors}. Results for the Sense of Self trials are in Table \ref{tab:empathy_self}.
        \end{tablenotes}
    \end{threeparttable}
\end{sidewaystable}

\clearpage
\begingroup
\singlespacing
\sloppy
\printbibliography[heading=subbibliography]
\endgroup

  \end{refsection}
\end{appendix}
\endgroup

\end{document}